\documentclass[10pt,twocolumn,letterpaper]{article}

\usepackage[pagenumbers]{cvpr} 

\usepackage{graphicx}
\usepackage{amsmath}
\usepackage{amssymb}
\usepackage{booktabs}

\usepackage{multirow}
\usepackage{array}
\usepackage{color}
\usepackage{xcolor}
\usepackage{bbm}
\usepackage{subcaption}
\usepackage{pifont}
\newcommand{\cmark}{\ding{51}}%
\newcommand{\xmark}{\ding{55}}%
\usepackage{makecell, booktabs}

\usepackage[pagebackref,breaklinks,colorlinks]{hyperref}

\usepackage[capitalize]{cleveref}
\crefname{section}{Sec.}{Secs.}
\Crefname{section}{Section}{Sections}
\Crefname{table}{Table}{Tables}
\crefname{table}{Tab.}{Tabs.}

\newcommand{\methodname}{\textsc{EnvEdit}}

\def\cvprPaperID{9403} 
\def\confName{CVPR}
\def\confYear{2022}

\begin{document}

\title{\methodname{}: Environment Editing for Vision-and-Language Navigation}

\author{Jialu Li \quad \quad Hao Tan \quad \quad Mohit Bansal
 \\ 
   UNC Chapel Hill\\ 
   \texttt{\{jialuli, airsplay, mbansal\}@cs.unc.edu}
}
\maketitle

\begin{abstract}
In Vision-and-Language Navigation (VLN), an agent needs to navigate through the environment based on natural language instructions. Due to limited available data for agent training and finite diversity in navigation environments, it is challenging for the agent to generalize to new, unseen environments. 
To address this problem, we propose \methodname{}, a data augmentation method that creates new environments by editing existing environments, which are used to train a more generalizable agent. 
Our augmented environments can differ from the seen environments in three diverse aspects: style, object appearance, and object classes. Training on these edit-augmented environments prevents the agent from overfitting to existing environments and helps generalize better to new, unseen environments. Empirically, on both the Room-to-Room and the multi-lingual Room-Across-Room datasets, we show that our proposed \methodname{} method gets significant improvements in all metrics on both pre-trained and non-pre-trained VLN agents, and achieves the new state-of-the-art on the test leaderboard. We further ensemble the VLN agents augmented on different edited environments and show that these edit methods are complementary.\footnote{Code and data are available at \url{https://github.com/jialuli-luka/EnvEdit}.}
\end{abstract}


\section{Introduction}

\begin{figure}[t]
\begin{center}
\includegraphics[width=1.0\linewidth]{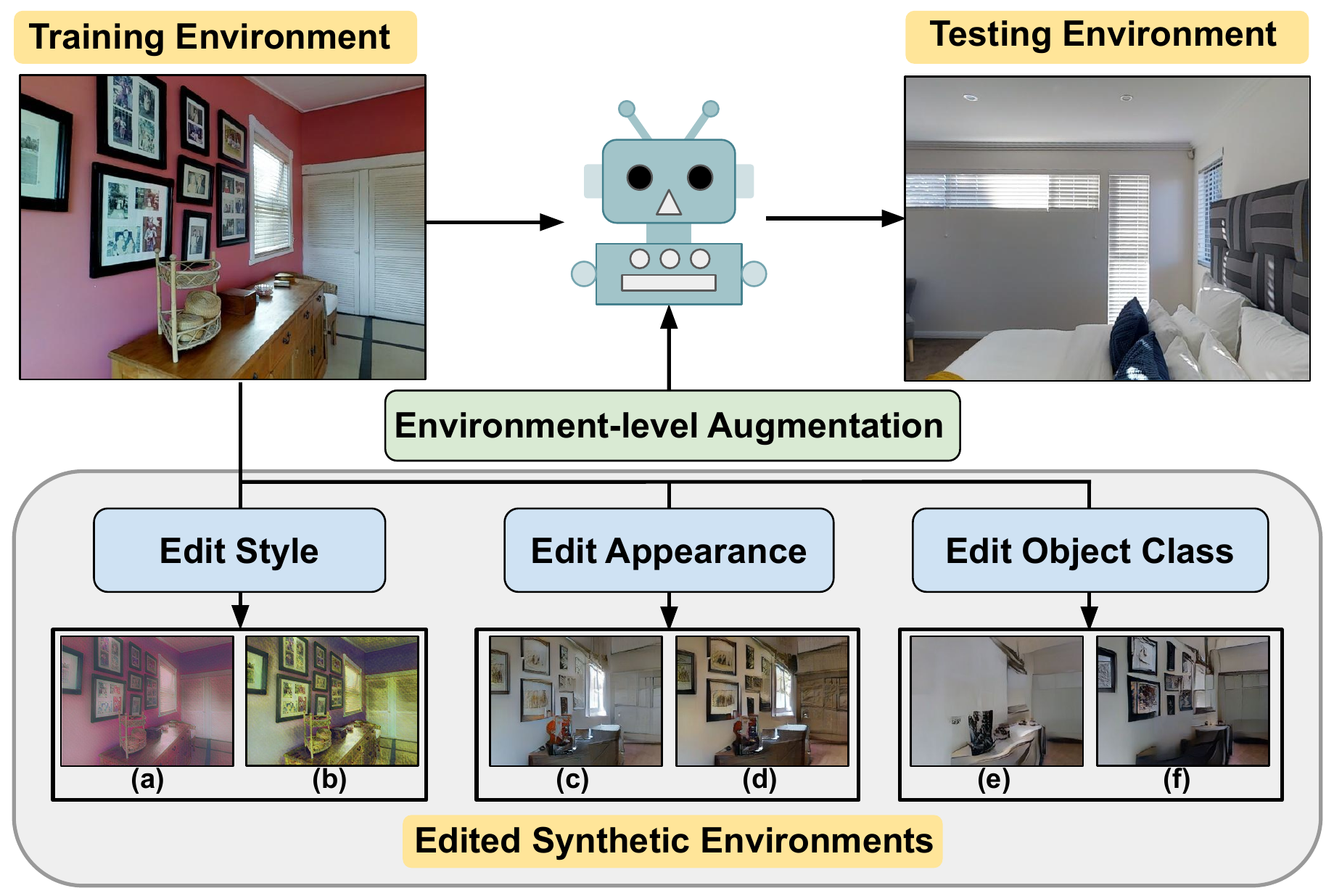}
\end{center}
\vspace{-12pt}
  \caption{We create synthetic environments through editing the style (a, b), object appearance (c, d), and object classes (e, f) of the training environment. Our synthetic environments serve as environment-level data augmentation during training and help the agent's generalization to the unseen test environment.
}
\label{figure1}
\vspace{-10pt}
\end{figure}

The Vision-and-Language Navigation (VLN) task requires an agent to navigate through the environment based on natural language instructions. Existing Vision-and-Language Navigation datasets are usually small in scale and contain a limited number of environments due to the difficulty of such data collection. However, navigation environments might differ greatly from each other. For example, indoor navigation environments might differ in the style of the room, the layout of the furniture, and the structure of the entire house. This makes it difficult for the agent to generalize to previously unseen environments. Previous works \cite{wang2019reinforced, fried2018speaker, ma2019self, zhu2020vision, huang2019transferable, hong2020language, kim-etal-2021-ndh} have seen that agents perform substantially worse in unseen environments, and many thoughtful methods \cite{tan2019learning, zhang2020diagnosing, wang2020environment, majumdar2020improving, guhur2021airbert, hong2020recurrent, liu2021vision} have been proposed to solve this generalization problem. 
One line of the previous work focuses on augmenting the environments to mitigate the environment bias. 
For example, \cite{tan2019learning} proposes to drop out environment-level features during training. 
However, this feature-dropping approach lacks the interpretability of the actually modified environments that the agent learns from to gain better generalizability. 
\cite{liu2021vision} takes one step further in creating in-domain augmentation data by mixing up existing training environments, which effectively reduces generalization error of VLN agents.
However, these mixed-up environments do not bring unseen changes or modifications to existing environments, and hence do not break the limitations of existing seen environments, which restricts agent's generalizability to unseen environments. 
Thus, in this paper, we propose to create new environments that differ from the original environments in style, appearance, and objects with style transfer and image synthesis approaches.  

Another line of work tries to address environment bias by pre-training from large image-text datasets \cite{majumdar2020improving, guhur2021airbert}, which equips the agent with diverse visual knowledge. 
Although promising performance was achieved, even pre-training data in \cite{guhur2021airbert}, which is an indoor room environment with captions collected from AirBnB, still differs from the VLN task in two ways.
In Vision-and-Language Navigation, the agent perceives a panoramic view and receives human-written language instructions, where in \cite{guhur2021airbert}, the panoramic view is the concatenation of images with similar semantics, and the instruction is a template-based mixing of image descriptions. 
This leads to a domain shift in pre-training data and might not adapt well to the VLN task.
Considering the large amount of pre-training data used, the performance gain on the Vision-and-Language Navigation task is still limited. 

To address these challenges, in this work, we propose \methodname{}: \textbf{Env}ironment \textbf{Edit}ing for Vision-and-Language Navigation. Our approach consists of three stages. In the first stage, we create new environments that maintain most of the semantic information of the original environments while changing the style, appearance, and object classes of the original environment. 
This constraint enables us to directly adopt the original human-annotated language instructions for the new environments, and avoid generating low-quality synthetic instructions \cite{zhao2021evaluation}. 
As illustrated in Figure~\ref{figure1},
our generated synthetic environments are mostly consistent with the original environments in semantics, but differ greatly in other aspects. 
For example, the overall style in Figure~\ref{figure1}~(a, b)  and object appearance in Figure~\ref{figure1}~(c, d)  are different, but the semantics of the synthetic environments mostly match the original environments. 
Meanwhile, our synthetic environments can also moderately differ from the original environments in object semantics (e.g., Figure~\ref{figure1}~(e) removes the pictures from the wall). 
Learning from these synthetic environments could enable the agent to better understand visual semantics and be more robust to appearance changes of objects in different environments.
Specifically, we adopt methods from style transfer \cite{jackson2019style} and image synthesis \cite{park2019semantic} to create new environments. 
In style transfer, the newly transferred environment is created with style embedding sampled from the learned embedding distribution of artistic paintings.
In image synthesis, we generate new environments based on semantic segmentation of the original environments, which changes the appearance of the objects. 
We further moderately edit the environment semantics and change objects (e.g., remove a lamp from the environment) by randomly masking some semantic classes in the semantic segmentation. 
In the second stage, the agent learns to navigate given natural language instructions from both the original environment and our aforementioned augmented environments. 
In the last stage, we follow the existing instruction-level data augmentation setup in \cite{fried2018speaker, tan2019learning}, which uses a speaker to generate new instructions for unannotated paths to fine-tune the agent. 
But different from \cite{tan2019learning}, our speaker is aware of styles and can generate different instructions given the style of the environment. 

We conduct experiments on both Room-to-Room (R2R) dataset \cite{anderson2018vision} and the multi-lingual Room-Across-Room (RxR) dataset \cite{ku2020room}. Empirical results show that our proposed \methodname{} outperforms all other non-pre-training methods by 1.6\% in success rate (SR) and 1.4\% in success rate weighted by path length (SPL) on R2R test leaderboard, and 5.3\% in normalized Dynamic Time Warping (nDTW) and 8.0\% in success rate weighted by normalized Dynamic Time Warping (sDTW) on RxR test leaderboard. 
We further show that our proposed approach is beneficial to SotA pre-trained agents. 
Our \methodname{} improves the performance by 3.2\% in SR and 3.9\% in SPL on R2R test leaderboard, and 4.7\% in nDTW and 6.6\% in sDTW on RxR test leaderboard, achieving the new state-of-the-art for both datasets. Lastly, we ensemble the VLN agents augmented on different edited environments and show that these editing methods are complementary to each other.

\begin{figure*}[t]
\begin{center}
\includegraphics[width=1.0\linewidth]{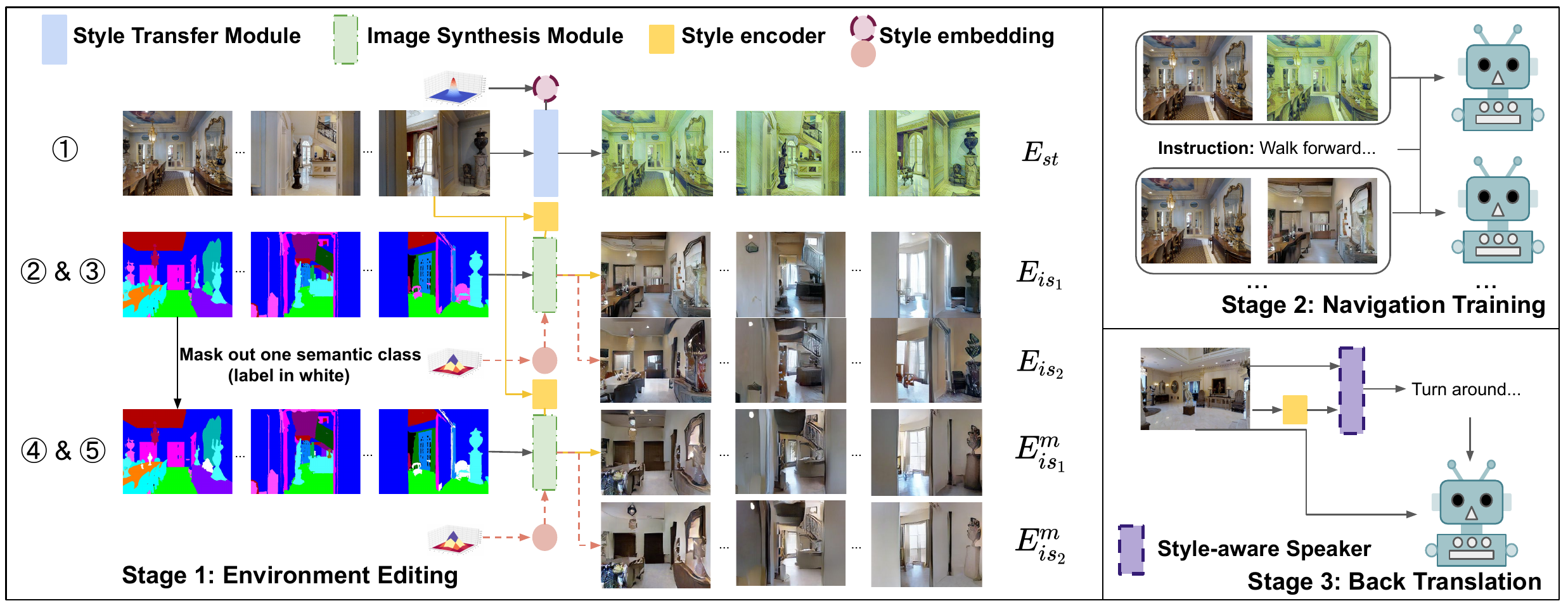}
\end{center}
\vspace{-10pt}
  \caption{Overview of our \methodname{}. In the first stage, the agent edits the original environment in five ways with style transfer and image synthesis approaches (Sec.~\ref{sec:gen}). Then, the agent is trained on Vision-and-Language Navigation task with both the original environment and the created environments (Sec.~\ref{sec:learning}). Lastly, a style-aware speaker is utilized to generate synthetic instructions for unannotated paths for back translation (Sec.~\ref{sec:learning}).
}
\label{figure2}
\vspace{-10pt}
\end{figure*}

\section{Related Work}
\paragraph{Vision-and-Language Navigation.} A lot of task setups, datasets, and simulators have been proposed for Vision-and-Language Navigation (VLN) \cite{nguyen2019help, anderson2018vision, chen2019touchdown, jain2019stay, shridhar2020alfred, berg2020grounding, hermann2020learning, misra2018mapping, mei2016listen, thomason2020vision}. In this paper, we focus on the Room-to-Room dataset \cite{anderson2018vision} and Room-Across-Room dataset \cite{ku2020room}, which have human-annotated instructions in different languages and simulated environments captured in Matterport3D \cite{chang2017matterport3d}. To solve this challenging task, a base agent contains cross-modality attention modules for cross-modal alignment between instructions and visual environments, LSTM \cite{hochreiter1997long} and transformer \cite{vaswani2017attention} based network to model context history and decode the sequence of navigation actions, and uses a mixture of reinforcement learning and imitation learning to train the agent \cite{wang2019reinforced, tan2019learning, ma2019self, li2021improving, landi2019perceive, zhu2021soon, DBLP:conf/acl/ZhuHCDJIS20}. In this paper, we build our methods on the strong baseline model EnvDrop \cite{tan2019learning} and further show our methods' compatibility to SotA pre-trained models \cite{hong2020recurrent, chen2021history}.

\paragraph{Mitigating Environment Bias.} Generalization to unseen environments is a key challenge in Vision-and-Language Navigation, especially for real-world environments. Many works have been proposed to mitigate the environment bias and enhance the performance in unseen environments \cite{tan2019learning, zhang2020diagnosing, wang2020environment, majumdar2020improving, guhur2021airbert, hong2020recurrent, liu2021vision}. One line of work focuses on feature-level engineering \cite{tan2019learning, wang2020environment}. 
However, these methods lack interpretability of what new environments the agent actually perceives and what semantic information across environments is learned by the model. 
Another line of work focuses on pre-training on large amounts of image-text pairs from other resources (e.g., web, image-caption datasets) or adopting pre-trained weights from SotA Vision-and-Language transformers \cite{li2020oscar, lu2019vilbert} to inject common sense visual and text knowledge into the model for better generalizability to unseen environments \cite{majumdar2020improving, guhur2021airbert, hong2020recurrent, moudgil2021soat, chen2021history}. 
Though the pre-training methods show promising performance, the pre-training data still differs from the panorama observations and human-annotated instructions in Vision-and-Language Navigation. To address the domain-shift between pre-training data and VLN data, we propose to use style transfer and image synthesis methods to augment the existing VLN data with new environments.

\paragraph{Data Augmentation in Vision-and-Language Navigation.} Data collection for Vision-and-Language Navigation task is resource consuming. Previous work in data augmentation mainly focuses on augmenting the instructions. \cite{fried2018speaker, tan2019learning} propose to train a speaker that generates instructions given unannotated paths from the seen environments, and \cite{zhu2020multimodal} proposes to transfer the style of the text for instruction augmentation. However, how to augment the training environment for better generalization is still underexplored. \cite{liu2021vision} proposes a useful environment-mixing method that creates new paths by mixing sub-paths from different environments. However, their approach is still limited by the existing seen environments, since they only concatenate existing environments and do not create environments that are new in style and unseen to the agent. \cite{koh2021pathdreamer} first tries to synthesize existing environments and predict future scenes for Vision-and-Language Navigation. 
In contrast to them, we propose \methodname{} that creates new environments for data augmentation and agents' generalization via style transfer and image synthesis methods. 

\paragraph{Data Augmentation in Computer Vision.} Data augmentation is a widely applied technique in the field of Computer Vision. Traditional data augmentation methods include random cropping, resizing, scaling, rotating, noise injection, image mixing, etc. \cite{lecun1998gradient, moreno2018forward, inoue2018data}. With advances in deep neural networks, using GAN \cite{goodfellow2014generative} based methods for data augmentation becomes popular \cite{bowles2018gan, perez2017effectiveness, zhu2018emotion, jackson2019style, shin2018medical, zheng2019stada, geirhos2018imagenet}. Following this trend, we use style transfer \cite{ghiasi2017exploring} and image synthesis \cite{park2019semantic} to create new environments for data augmentation.

\section{Method Overview}

\subsection{Problem Setup} \label{sec:problem}
Vision-and-Language Navigation (VLN) requires an agent to navigate through the environment based on natural language instructions. Formally, given a natural language instruction $I$, at each time step $t$, the agent perceives a panoramic view $P_t$ of the current location, and needs to pick the next viewpoint from a set of $K$ navigable locations $\{g_{t,k}\}_{k=1}^{K}$. Specifically, the panoramic view $P_t$ is discretized into 36 single views $\{p_{t,i}\}_{i=1}^{36}$. Each view representation $f_{t,i}$ is the concatenation of its visual representation $v_{t,i}$ encoded by the pre-trained vision model and its orientation feature $o_{t,i} = (\cos\theta_{t,i}, \sin\theta_{t,i}, \cos\phi_{t,i}, \sin\phi_{t,i})$ that encodes heading $\theta_{t,i}$ and elevation $\phi_{t,i}$ information. The navigable location is represented as the visual feature of one specific view that is closest to the navigable direction from 36 discretized views and its orientation feature. The agent will predict a ``STOP" action when the navigation ends.

\subsection{Training Procedures} \label{sec:learning}

The overview of our \textbf{Env}ironment \textbf{Edit}ing (\methodname{}) approach is shown in Figure~\ref{figure2}. It contains three stages. We describe the three stages briefly in this section.

\paragraph{Environment Creation.} In the first stage, we create multiple environments that differ from the original environments in style, appearance, and object classes (described in Sec~\ref{sec:st} and Sec~\ref{sec:is}). We adopt the off-the-shelf models from \cite{jackson2019style} for style transfer. For image synthesis, we train the image generator and style encoder on all the seen environments in Room-to-Room dataset \cite{anderson2018vision}. 

\paragraph{Vision-and-Language Navigation Training.} In the second stage, the agent is trained on both the original environments and the new environments on the Vision-and-Language Navigation task. Specifically, in a batch of $N$ instruction-path pairs, half of the pairs will observe the original environments, and the other half will perceive the edited environments. This prevents the agent from overfitting to the original environments. A mixture of imitation learning and reinforcement learning is adopted as in \cite{tan2019learning}.

\paragraph{Back Translation.} In the third stage, we follow \cite{tan2019learning} to do back translation, which generates synthetic instructions for unannotated paths from seen environments with a speaker. The agent is trained on both the original and the newly generated instruction-path pairs. The speaker used in \cite{tan2019learning} consists a two-layer bi-directional LSTM \cite{hochreiter1997long} that encodes route information $\{p_i\}_{i=1}^{L}$ and context information $\{c_i\}_{i=1}^{L}$ hierarchically, and a traditional LSTM based decoder with attention over encoded context information to generate synthetic instructions. To better serve the environment creation purpose, we enhance this speaker by further incorporating style information of the route.
Specifically, we initialize the speaker decoder with the style embedding of the start viewpoint on the route:

\begin{align}
    x_0 &= \mathrm{LSTM}(w_0, (h_\mathit{style}, c_\mathit{style})) \\ 
    (h_\mathit{style}, c_\mathit{style}) &= \mathrm{FC Layer}(s_0)  \\ 
    s_0 &= \frac{1}{36}\sum_{k=1}^{36}\mathrm{StyleEncoder}(o_{0,k})
\end{align}
where $o_{0,k}$ is the discretized view for the start location and $w_{0}$ is the start token for instruction generation. $x_0$ is further attended with context information $\{c_i\}_{i=1}^{L}$ to predict the next word in the instruction. 

\section{Environment Editing} \label{sec:gen}
In this section, we describe the environment editing methods that we use to create new environments. We focus on editing three components $(S, A, O)$ of an environment $E$, where $S$ is the style of the environment, $A$ is the object appearance, and $O$ is the class of objects (indicated by the semantic segmentation mask of the environment).
In Sec~\ref{sec:st}, we present the style transfer we use to edit the style $S$ of an environment. In Sec~\ref{sec:is}, we use image synthesis to edit the object appearance $A$ of an environment, while also providing the option to edit the style $S$ and the objects $O$. An example of different kinds of created environments is shown in Figure~\ref{figure2} Stage 1.

\subsection{Style Transfer} \label{sec:st}
Previous environment-level data augmentation methods in Vision-and-Language Navigation mainly focus on feature augmentation (i.e., adding random noise directly to the visual representation encoded with pre-trained vision models) \cite{tan2019learning} and environment mixing augmentation (i.e., mixing up paths from two training environments) \cite{liu2021vision}. Though both methods are useful and achieve promising results, feature augmentation has the issue of hard to interpret, and environment mixing augmentation does not address the limitations of existing environments and sometimes the mixed scenes are unrealistic (e.g., navigating from a modern living house to a museum-style room).
To address these issues, we propose to create new environments that are semantically consistent with the original environments but different in style. Our created new environments could potentially mimic the unseen environments, and are more realistic compared with \cite{liu2021vision}. The main advantage of maintaining the semantics of the original environments is that the original human-annotated language instructions can be directly adapted to the new environments with high correspondence. 
This eliminates the need to generate synthetic instructions \cite{tan2019learning, fried2018speaker} for new environments, which has shown to be much worse than human annotations \cite{zhao2021evaluation}.
The new environment $E_{st}$ ($st$ -- \textbf{S}tyle \textbf{T}ransfer) we create is $(S_{st}, A_o, O_o)$, which differs from the original environment $E_o$ in style of the environment. Specifically, we follow the approach \cite{jackson2019style} that is both computationally efficient and has high quality output.

\paragraph{Style Transfer Model Architecture.}  
The architecture of the style transfer approach we use is shown in Figure~\ref{figure2} Stage 1 row \textcircled{\raisebox{-0.9pt}{1}}. The content image is encoded by a CNN based architecture with residual connections. The style embedding is sampled from a multivariate normal distribution, whose mean and covariance is from the distribution of the style embeddings from the Painter By Numbers (PBN) dataset\footnote{https://www.kaggle.com/c/painter-by-numbers}. During decoding, the sampled style embedding is incorporated by the conditional instance normalization \cite{dumoulin2016learned}:
\begin{equation}
    x_\mathit{out} = \gamma_\mathit{style}(\frac{x_\mathit{in}-\mu_\mathit{in}}{\sigma_\mathit{in}}) + \beta_\mathit{style}
\end{equation}
where $\gamma_\mathit{style}$ and $\beta_\mathit{style}$ are computed by passing the sampled style embedding through two separate fully-connected layers, and $\mu_\mathit{in}$ and $\sigma_\mathit{in}$ are the mean and standard deviation of the encoded content image $x_\mathit{in}$ respectively.

\paragraph{Fixed style for discretized views.} 
At each time step, the agent perceives a panoramic view $P_t$ of the current location, which is discretized into 36 single views $\{p_{t,i}\}_{i=1}^{36}$. 
The 36 single views are correlated to each other, and there exists overlaps between adjacent views. Thus, to keep the visual observation consistent in style in a panoramic view at one time step, we sample the same style embedding from the multivariate normal distribution for all the 36 discretized views. We show in Sec~\ref{sec:fix} that this setup is crucial for creating effective edited environments for the agent.

\subsection{Image Synthesis} \label{sec:is}
The style transfer approach creates environments that maintain the semantics of the original environments and only change the style. Nevertheless, the appearance and texture of objects in the environment remain the same.  
Thus, we explore one step further by creating new environments that are semantically similar to the original environments but different in both the style and the appearance of the objects with the image synthesis approach. We explore a specific form of conditional image synthesis, which generates a new photorealistic image conditioned on a semantic segmentation mask. In this setup, the semantics of the new environments $E_{is}$ are constrained by the semantic segmentation of the original environments, while the shape and appearance $A$ of the objects can be diversely generated by the model. We further explore generating environments that have different objects $O$ by changing one of the semantic classes in the semantic segmentation. In both cases, the synthetic environments have high correspondence with the original instructions, since the semantics remain unchanged or only slightly different from the original environments. 
We adopt the approach from \cite{park2019semantic} for semantic image synthesis. 

\paragraph{Model Architecture.} 
The image generator is a GAN based conditional image synthesis model as in \cite{park2019semantic}. Specifically, the model contains several ResNet blocks with upsampling layers. SPADE blocks \cite{park2019semantic} are used to learn the parameters for normalization layers and are conditioned on semantic segmentation mask information. The model is trained on GAN hinge loss \cite{lim2017geometric} and feature matching loss \cite{wang2018high}. 
Given the semantic segmentation mask, we could control the style of the synthesis image by using different style embeddings as input to the generator. Following \cite{park2019semantic}, we learn an encoder that maps a style image to a style embedding by adding a KL-Divergence loss during training.

\paragraph{Editing Appearance.}
With the image generator and style encoder, we create two kinds of environments that edit the appearance of the original environments (shown in Figure~\ref{figure2} Stage 1 row \textcircled{\raisebox{-0.9pt}{2}} and \textcircled{\raisebox{-0.9pt}{3}}). The first kind of environment is $E_{is_1}$ ($is$ -- \textbf{I}mage \textbf{S}ynthesis) with components $(S_o, A_{is_1}, O_o)$, which differs from the original environments only in object appearance. We create this kind of environment by using the views in the original environments as the style image to maintain the style of the original environments. The second kind of environment is $E_{is_2}$ with components $(S_{is_2}, A_{is_2}, O_o)$, created by manually setting a fixed style embedding (e.g., all-zero embedding) for the generator. This new environment differs from the original environments $E_o$ in both style and object appearance. With these two kinds of environments, we are able to explore the impact of the style $S$ and object appearance $A$ separately.

\paragraph{Editing Objects.} 
After creating environments with different styles $S$ and object appearances $A$, we take one step further to remove and change some of the objects in the original environments (shown in Figure~\ref{figure2} Stage 1 row \textcircled{\raisebox{-0.9pt}{4}} and \textcircled{\raisebox{-0.9pt}{5}}). Though there exists many works in (text-guided) image manipulation \cite{liu2020describe, li2020manigan, chen2018language, cheng2020sequential, dong2017semantic, el2019tell, nam2018text, shinagawa2018interactive, meng2021sdedit, bau2021paint}, for simplicity, we change the objects by modifying the semantic segmentation of the original environments. Specifically, suppose that the original semantic segmentation contains $C$ classes, we add a ``mask" class and use it as the $C+1$ class. During training, we randomly pick one class from the $C$ classes and set it to be the ``mask" class. In this case, the model
could generate random masks for the ``mask" class. We create new environments $E_{is_1}^m(S_o, A_{is_1}^m, O_{is_1}^m)$ and $E_{is_2}^m(S_{is_2}^m, A_{is_2}^m, O_{is_2}^m)$ ($E_{is}^m$ -- \textbf{I}mage \textbf{S}ynthesis with \textbf{M}asks) by randomly masking out objects from the original environment $E_o$. $E_{is_1}^m$ and $E_{is_2}^m$ differ in that $E_{is_1}^m$ maintains the style of the original environment, while $E_{is_2}^m$ changes all three components of the original environment $E_o$. The style changes are controlled by the style embedding from the style encoder.

\begin{table*}
    \centering
    \resizebox{1.8\columnwidth}{!}{
    \begin{tabular}{|c|ccc|cccc|cccc|}
    \hline 
        & \multicolumn{3}{c|}{\textbf{Environment Components}}  & \multicolumn{4}{c|}{\textbf{ViT-B/32}} & 
       \multicolumn{4}{c|}{\textbf{ViT-B/16}}\\  \hline 
       \textbf{Models} & \textbf{Style} & \textbf{Appearance} & \textbf{Object} & \textbf{TL} & \textbf{NE$\downarrow$} & \textbf{SR$\uparrow$} & \textbf{SPL$\uparrow$}   & \textbf{TL} & \textbf{NE$\downarrow$} & \textbf{SR$\uparrow$} & \textbf{SPL$\uparrow$}  \\ \hline
EnvDrop$^*$ \cite{shen2021much} & \xmark & \xmark & \xmark & 14.339 & 5.214 & 51.3 & 45.8 & 15.861 & 4.734 & 55.1 & 48.8 \\ \hline 
$E_{st}$ & \cmark & \xmark & \xmark & 14.738 & \textbf{4.631} & \textbf{56.5} & \textbf{50.7} & 16.585 & 4.690 & \textbf{58.2} & \textbf{51.5} \\
$E_{is_1}$ & \xmark & \cmark & \xmark & 15.871 & 4.766 & 56.2 & 49.8 & 17.690 & 4.759 & 56.4 & 48.9 \\ 
$E_{is_2}$ & \cmark & \cmark & \xmark  & 15.427 & 5.049 & 54.2 & 48.5 & 15.273 & 4.767 & 56.2 & 49.6 \\ 
$E_{is_1}^m$ & \xmark & \cmark & \cmark & 15.788 & 4.966 & 54.2 & 49.7 & 14.464 & 4.666 & 57.3 & 51.1 \\
$E_{is_2}^m$ & \cmark & \cmark & \cmark & 17.906 & 4.979 & 54.2 & 47.6 & 14.204 & \textbf{4.607} & 56.1 & 50.8 \\
    \hline
    \end{tabular}
    }
    \caption{Performance of training the agent with one kind of our edited environments. Results are on R2R val-unseen set. ViT-B/32(16) indicate image features extracted with different CLIP-ViT models \cite{radford2021learning}. ``*" indicates reproduced results. \cmark \  indicates the environment component of the new environment is different from the original environment, while \xmark \  indicates the same.
    }
    \label{table1}
\end{table*}

\section{Experimental Setup}

\subsection{Datasets}
We evaluate our agent on the Room-to-Room (R2R) dataset \cite{anderson2018vision} and the Room-Across-Room (RxR) dataset \cite{ku2020room}. R2R dataset contains English instructions and RxR dataset contains instructions in English, Hindi, and Telugu. Both datasets are split into a training set, seen and unseen validation sets, and a test set. The environments in the unseen validation set and test set do not appear in the training set. 

\subsection{Evaluation Metrics}
We evaluate our model with six metrics: (1) Success Rate (SR). (2) Success Rate weighted by Path Length (SPL) \cite{anderson2018evaluation}. (3) Trajectory Length (TL). (4) Navigation Error (NE). (5) normalized Dynamic Time Warping (nDTW) \cite{ilharco2019general}. (6) success rate weighted by normalized Dynamic Time Warping (sDTW) \cite{ilharco2019general}. SR, SPL are the main metrics for evaluation on R2R dataset, and nDTW, sDTW are the main metrics for RxR dataset. Details can be found in Appendix.

\section{Results and Analysis}
In this section, we first compare the performance of training on different environments that we created in Sec.~\ref{sec:env}. Then, we show that our method could generalize to pre-trained navigation agents in Sec.~\ref{sec:rec}. We further show the importance of using a fixed style for discretized views and our style-aware speaker through ablations in Sec.~\ref{sec:fix}. Moreover, we demonstrate that our editing methods are complementary to each other in Sec.~\ref{sec:ensemble}. Lastly, we show our model's performance on the test leaderboards of both Room-to-Room dataset and Room-Across-Room dataset in Sec.~\ref{sec:test}. We demonstrate some qualitative examples of our edited environments in Sec.~\ref{sec:qual}.

\subsection{Results for Environment Editing Methods} \label{sec:env}
In the environment creation stage, we create five kinds of environments that differ in style, appearance and objects $E_{st}, E_{is_1}, E_{is_2}, E_{is_1}^m, E_{is_2}^m$ with editing methods as described in Sec~\ref{sec:st} and Sec~\ref{sec:is}. We show the performance of training with the original environments and one of the new environments on R2R dataset in Table~\ref{table1}. Back translation is not applied in these experiments, and could be found in Appendix.
We can see that training with any of the newly created environments can outperform the baseline model by a large margin on the validation unseen set. 
Specifically, the $E_{st}$ environment, which differs from the original environment in style only, achieves the best performance, improving the baseline trained on ViT-B-32 features by 5.2\% in SR and 4.9\% in SPL, and a stronger baseline trained on ViT-B-16 features by 3.1\% in SR and 2.7\% in SPL. This demonstrates that augmenting the training environment with synthetic new environments helps generalization to unseen data, regardless of the visual features.

Overall, comparing the environments created with style transfer approach $E_{st}$ and the environments created with image synthesis approaches \{$E_{is_1}$, $E_{is_2}$, $E_{is_1}^m$, $E_{is_2}^m$\}, $E_{st}$ brings slightly higher improvement in both SR and SPL. We attribute this to the higher environment creation quality of the style transfer approach compared with the image synthesis approach. 
Comparing $E_{is_1}$ and $E_{is_2}$, it shows that keeping the style unchanged while modifying the appearance of the objects improves the SR by 2.0\% for model trained with ViT-B/32 features and 0.2\% for model trained with ViT-B/16 features. This is because that the new environments that maintain the style will have a higher correspondence with the original instructions, while also being different enough. Given that we do not generate synthetic instructions for new environments (due to the low quality of synthetic instructions  \cite{zhao2021evaluation}), it is important to find a balance between the matches to the original instructions and the diversity of the new environments. 
Similar results are observed for $E_{is_1}^m$, $E_{is_2}^m$. 

Furthermore, comparing \{$E_{is_1}$, $E_{is_1}^m$\}, we observe that features that learned from smaller patches (ViT-B/16) could benefit from the slight removal or change of objects in the environments. Similar performance improvement is observed for pre-trained VLN agent with ViT-B/16 features (discussed in Sec.~\ref{sec:rec}).

Lastly, we observe that with different visual backbones (ViT-B/32 and ViT-B/16), and different VLN base models (discussed in Sec.~\ref{sec:rec}), the improvement brought by different synthetic environments are inconsistent. For example, with the same base agent EnvDrop, training on $E_{is_1}$ works better than training on $E_{is_1}^m$ for ViT-B/32 features, but not for ViT-B/16 features. We attribute this to features extracted with different visual backbones generalize to unseen environments differently. Detailed analysis can be found in the Appendix. Considering both simplicity and performance across different visual backbones and VLN base models, we recommend using $E_{st}$ as a start point in future research.

\begin{table}
    \centering
    \resizebox{0.7\columnwidth}{!}{
    \begin{tabular}{ccccc}
    \hline 
      \textbf{Models}   &  \textbf{TL} & \textbf{NE$\downarrow$} & \textbf{SR$\uparrow$} & \textbf{SPL$\uparrow$} \\ \hline
        HAMT\cite{chen2021history} & - & - & 65.7 & 60.9   \\
        $E_{st}$-16 & 11.78 & 3.42 & 67.3 & 62.6 \\
        $E_{is_1}$-16 & 11.23 & 3.52 & 66.8 & 62.1 \\ 
        $E_{is_1}^m$-16 &  12.13 & \textbf{3.22} & \textbf{67.9} & \textbf{62.9}  \\ \hline 
    \end{tabular}
    }
    \caption{Performance of applying our proposed method to SotA VLN agents on R2R validation unseen set.}
    \label{table2}
\end{table}

\subsection{Performance on Pre-trained VLN Agents} \label{sec:rec}
In this section, we show that our \methodname{} is complementary to the VLN pre-training methods. We enhance the SotA pre-traind VLN model HAMT \cite{chen2021history} with our methods and illustrate the improvements on R2R dataset.

The model architecture of \cite{chen2021history} is based on transformer. The image feature used in this work is extracted with CLIP ViT-B/16 without the last linear representation layer. We follow their work to extract the visual features for our created environments, and directly fine-tune their released pre-trained models with \methodname{}.
As shown in Table~\ref{table2}, augmenting the original environment with $E_{is_1}^m$ could improve the baseline by 2.2\% in SR and 2.0\% in SPL. Augmenting with the other two environments could also improve the baseline by around 1.5\% in both SR and SPL. This demonstrates the effectiveness of adapting our method to strong SotA VLN models.

\subsection{Method Ablations} \label{sec:fix}
In this section, we show two ablations for our proposed method. We first show that using a fixed style for all 36 discretized views of a panorama is essential for creating new environments for the agent. Then, we show that our style-aware speaker achieves better performance when used in back translation compared with the baseline speaker.

\begin{table}
    \centering
    \resizebox{0.9\columnwidth}{!}{
    \begin{tabular}{ccccc}
    \hline 
      \textbf{Models}   &  \textbf{TL} & \textbf{NE$\downarrow$} & \textbf{SR$\uparrow$} & \textbf{SPL$\uparrow$} \\ \hline
      EnvDrop-16$^*$ \cite{shen2021much} & 15.86 & 4.73 & 55.1 & 48.8  \\
        $E_{st}$-16 & 16.59 & \textbf{4.69} & \textbf{58.2} & \textbf{51.5} \\
        $E_{st}$-16 w/o fixed views & 16.79 & 4.70 & 55.9 & 48.8 \\ 
        $E_{st}$-16 w/ fixed env & 14.36 & 4.70 & 55.8 & 49.5\\ \hline 
        EnvDrop-32$^*$ \cite{shen2021much} &  14.34 & 5.21 & 51.3 & 45.8 \\
        $E_{st}$-32 & 14.74 & \textbf{4.63} & \textbf{56.5} & \textbf{50.7} \\
        $E_{st}$-32 w/o fixed views & 14.50 & 4.88 & 54.7 & 48.8 \\  
        $E_{st}$-32 w/ fixed env & 17.39 & 4.87 & 55.0 & 48.5 \\ \hline 
    \end{tabular}
    }
    \caption{Ablation results on R2R val-unseen set illustrating the benefit of using the fixed style for a panorama. 
    ``-16" and ``-32" indicate image features extracted with ViT-B/16(32). ``*" indicates reproduced results.}
    \label{table3}
\end{table}

\vspace{2pt}
\noindent\textbf{Fixed style for discretized views.}
As shown in Table~\ref{table3}, when the style is different inside a panorama (``$E_{st}$-16" vs. ``$E_{st}$-16 w/o fixed views"), the performance drops by 2.3\% in SR and 1.7\% in SPL. This indicates that using a fixed style for 36 discretized views of a panorama is essential for the performance improvement, since it provides consistent visual semantics. Furthermore, we show that keeping a fixed style for the whole environment (``$E_{st}$-16 w/ fixed env") will decrease the improvements by 2.4\% in SR and 2.0\% in SPL. Similar results are observed for ViT-B/32 features.
This indicates that using a fixed style at each viewpoint has a better balance between consistency in observation and variance in style. 

\vspace{2pt}
\noindent\textbf{Style-aware Speaker.}
As shown in Table~\ref{table4}, our style-aware speaker improves the performance in SR and SPL by around 1\% for both features (ViT-B/32 and ViT-B/16). This implies that explicitly incorporating environment style helps generate synthetic instructions that match with the environments better. Besides, we show that our style-aware speaker can improve the overall performance for different kinds of created environments (i.e., $E_{st}$, $E_{is_1}^m$). 

\begin{table}
    \centering
    \resizebox{0.8\columnwidth}{!}{
    \begin{tabular}{ccccc}
    \hline 
      \textbf{Models}   &  \textbf{TL} & \textbf{NE$\downarrow$} & \textbf{SR$\uparrow$} & \textbf{SPL$\uparrow$} \\ \hline
      $E_{st}$-32 + BT  &  17.777 & 4.504 & 59.0 & 51.8 \\
       $E_{st}$-32 + BTS & 15.912 & \textbf{4.335} & \textbf{60.2} & \textbf{53.8}\\  \hline 
       $E_{is_1}^m$-16 + BT & 16.752 & 4.316 & 60.2 & 53.4 \\ 
       $E_{is_1}^m$-16 + BTS & 17.989 & \textbf{4.232} & \textbf{60.8} & \textbf{54.2} \\
    \hline
    \end{tabular}
    }
    \caption{Ablation results on R2R val-unseen set showing the improvement of our style-aware speaker. `+BT" indicates back translation with the baseline speaker, and ``+BTS" indicates using style-aware speaker in back translation. }
    \vspace{-5pt}
    \label{table4}
\end{table}

\begin{table}
    \centering
    \resizebox{0.9\columnwidth}{!}{
    \begin{tabular}{ccccc}
    \hline 
      \textbf{Models}   &  \textbf{TL} & \textbf{NE$\downarrow$} & \textbf{SR$\uparrow$} & \textbf{SPL$\uparrow$} \\ \hline
        $E_{st}$-ED & 16.59 & 4.69 & 58.2 & 51.5   \\
        $E_{st}$+$E_{is_1}$+$E_{is_1}^m$-ED & 15.60 & \textbf{4.52} & \textbf{58.8} & \textbf{52.7} \\ \hline 
        $E_{is_1}^m$-H & 12.13 & 3.22 & 67.9 & 62.9    \\
        $E_{st}$+$E_{is_1}$+$E_{is_1}^m$-H & 11.13 & \textbf{3.24} & \textbf{68.9} & \textbf{64.4}   \\ \hline
    \end{tabular}
    }
    \caption{Performance of ensembling VLN agents trained on different environments. ``ED" and ``H" indicates using EnvDrop and HAMT as the base navigation agents respectively. }
    \vspace{-5pt}
    \label{table_combine}
\end{table}

\begin{table*}
\begin{small}
\centering
\resizebox{1.8\columnwidth}{!}{
\begin{tabular}{p{0.32\columnwidth}|>{\centering\arraybackslash}p{0.09\columnwidth}>{\centering\arraybackslash}p{0.09\columnwidth}>{\centering\arraybackslash}p{0.09\columnwidth}>{\centering\arraybackslash}p{0.09\columnwidth}|>{\centering\arraybackslash}p{0.09\columnwidth}>{\centering\arraybackslash}p{0.09\columnwidth}>{\centering\arraybackslash}p{0.09\columnwidth}>{\centering\arraybackslash}p{0.09\columnwidth}|>{\centering\arraybackslash}p{0.09\columnwidth}>{\centering\arraybackslash}p{0.09\columnwidth}>{\centering\arraybackslash}p{0.09\columnwidth}>{\centering\arraybackslash}p{0.09\columnwidth}}
\hline 
\multicolumn{1}{c}{\textbf{Models}} & \multicolumn{4}{c}{\textbf{Validation Seen}} & \multicolumn{4}{c}{\textbf{Validation Unseen}} & \multicolumn{4}{c}{\textbf{Test Unseen}} \\ \hline
 & \textbf{TL} & \textbf{NE$\downarrow$} & \textbf{SR$\uparrow$} & \textbf{SPL$\uparrow$} & \textbf{TL} & \textbf{NE$\downarrow$} & \textbf{SR$\uparrow$} & \textbf{SPL$\uparrow$} & \textbf{TL} & \textbf{NE$\downarrow$} & \textbf{SR$\uparrow$} & \textbf{SPL$\uparrow$} \\ \hline
$\circlearrowright$BERT$^{\spadesuit}$ \cite{hong2020recurrent} & 11.13 & 2.90 & 72 & 68 & 12.01 &  3.93 & 63 & 57 & 12.35 & 4.09 & 63 & 57 \\ 
EnvDrop-CLIP \cite{shen2021much} & - & - & - & - & - & - &  59.2 & 52.9 & - & - & 59 & 53 \\ 
AirBERT$^{\spadesuit}$ \cite{guhur2021airbert} & 11.09 & 2.68 & 75 & 70 & 11.78 & 4.01 & 62 & 56 &  12.41 & 4.13 & 62 & 57 \\
HAMT$^{\spadesuit}$ \cite{chen2021history} &  11.15 &  2.51 & 76 & 72 & 11.46 &  \textbf{2.29} & 66 & 61 &  12.27 & 3.93 & 65 & 60 \\
REM$^{\spadesuit}$ \cite{liu2021vision} & 10.88 & 2.48 & 75.4 & 71.8 & 12.44 & 3.89 & 63.6 & 57.9 & 13.11 & 3.87 & 65.2 & 59.1 \\ 
Ours  & 14.64 & 3.35 & 69.4 & 64.2 & 17.99 & 4.23 & 60.8 & 54.2 & 16.84 & 4.30 & 60.6 & 54.4 \\ 
Ours$^{\spadesuit}$ & 11.18 & \textbf{2.32} & \textbf{76.9} & \textbf{73.9} & 11.13 & 3.24 & \textbf{68.9} & \textbf{64.4} & 11.90 & \textbf{3.59} & \textbf{68.2} & \textbf{63.9}  \\ \hline 
\end{tabular}
}
\caption{Comparison of agent performance on R2R dataset under the single-run setting. $\spadesuit$ indicates pre-trained VLN agents.}
\label{table5}
\end{small}
\end{table*}

\begin{table*}
\begin{small}
\centering
\resizebox{1.7\columnwidth}{!}{
\begin{tabular}{p{0.33\columnwidth}|>{\centering\arraybackslash}p{0.12\columnwidth}>{\centering\arraybackslash}p{0.12\columnwidth}>{\centering\arraybackslash}p{0.12\columnwidth}|>{\centering\arraybackslash}p{0.12\columnwidth}>{\centering\arraybackslash}p{0.12\columnwidth}>{\centering\arraybackslash}p{0.12\columnwidth}|>{\centering\arraybackslash}p{0.12\columnwidth}>{\centering\arraybackslash}p{0.12\columnwidth}>{\centering\arraybackslash}p{0.12\columnwidth}}
\hline 
\multicolumn{1}{c}{\textbf{Models}} & \multicolumn{3}{c}{\textbf{Validation Seen}} & \multicolumn{3}{c}{\textbf{Validation Unseen}} & \multicolumn{3}{c}{\textbf{Test Unseen}} \\ \hline
  & \textbf{SR$\uparrow$} & \textbf{NDTW$\uparrow$} & \textbf{SDTW$\uparrow$} & \textbf{SR$\uparrow$} & \textbf{NDTW$\uparrow$} & \textbf{SDTW$\uparrow$} & \textbf{SR$\uparrow$} & \textbf{NDTW$\uparrow$} & \textbf{SDTW$\uparrow$}  \\ \hline
EnvDrop-CLIP \cite{shen2021much} & - & - & - & 42.6 & 55.7 & - & 38.3 & 51.1 & 32.4 \\
HAMT$^{\spadesuit}$  \cite{chen2021history} & 59.4 & 65.3 & 50.9 & 56.5 &  63.1 & 48.3 & 53.1 &  59.9 & 45.2 \\
Ours & 53.1 & 63.0 & 45.9 & 50.1 & 60.6 & 43.0 & 46.2 & 56.4 & 40.4 \\ 
Ours$^{\spadesuit}$ & \textbf{67.2} & \textbf{71.1} & \textbf{58.5} & \textbf{62.8} & \textbf{68.5} & \textbf{54.6} & \textbf{60.4} & \textbf{64.6} & \textbf{51.8} \\ 
\hline
\end{tabular}
}
\caption{Comparison of agent performance on RxR dataset under the single-run setting.  $\spadesuit$ indicates pre-trained VLN agents.}
\vspace{-10pt}
\label{table6}
\end{small}
\end{table*}

\subsection{Combining Multiple Environments} \label{sec:ensemble}
In this section, we discuss our initial exploration for combining multiple environments, where we use the traditional ensemble method to boost overall performance. Specifically, the agent makes its decision based on the average logits predicted by all the ensembled models. As shown in Table~\ref{table_combine}, for agents that use EnvDrop as the base agent, simply ensembling the VLN agents trained on three edited environments ($E_{st}$, $E_{is_1}$, $E_{is_1}^m$) could slightly improve the overall performance by 0.6\% in SR and 1.2\% in SPL compared with augmenting only with $E_{st}$. Similar improvement is observed when using HAMT as the base agent. We further explore combining multiple environments during training using adaptive curriculum learning in the Appendix. 

\begin{table}[t]
    \centering
    \resizebox{0.95\columnwidth}{!}{
    \begin{tabular}{cccc}
\hline 
\multicolumn{1}{c}{\textbf{Original}} &
\multicolumn{1}{c}{$E_{st}$} & 
\multicolumn{1}{c}{$E_{is_1}$} & 
\multicolumn{1}{c}{$E_{is_1}^m$} \\ \hline 
\begin{minipage}{.2\columnwidth}
     \includegraphics[width=\columnwidth]{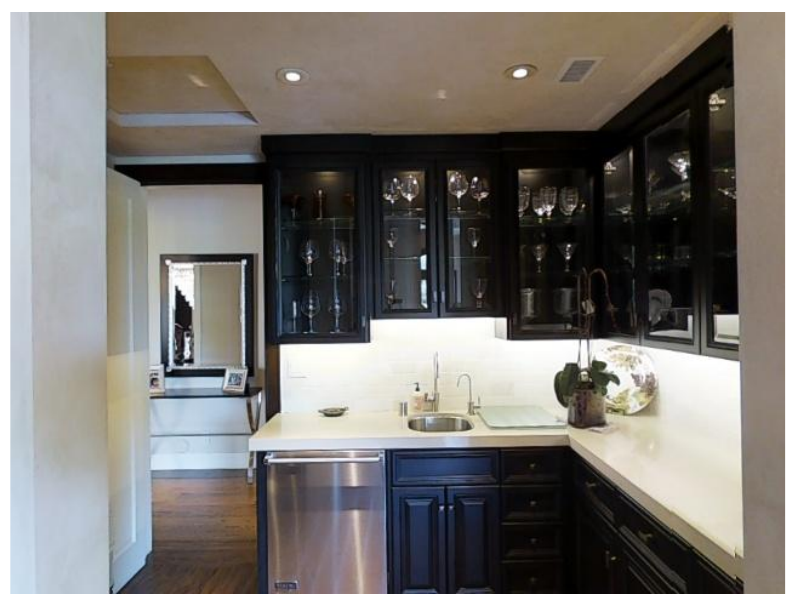} 
    \end{minipage} & \begin{minipage}{.2\columnwidth}
      \includegraphics[width=\columnwidth]{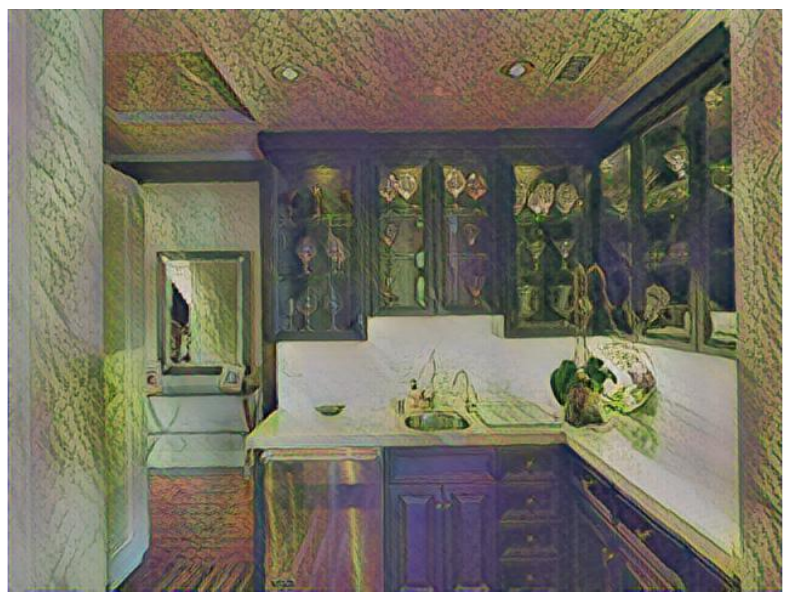} 
    \end{minipage} & \begin{minipage}{.2\columnwidth}
      \includegraphics[width=\columnwidth]{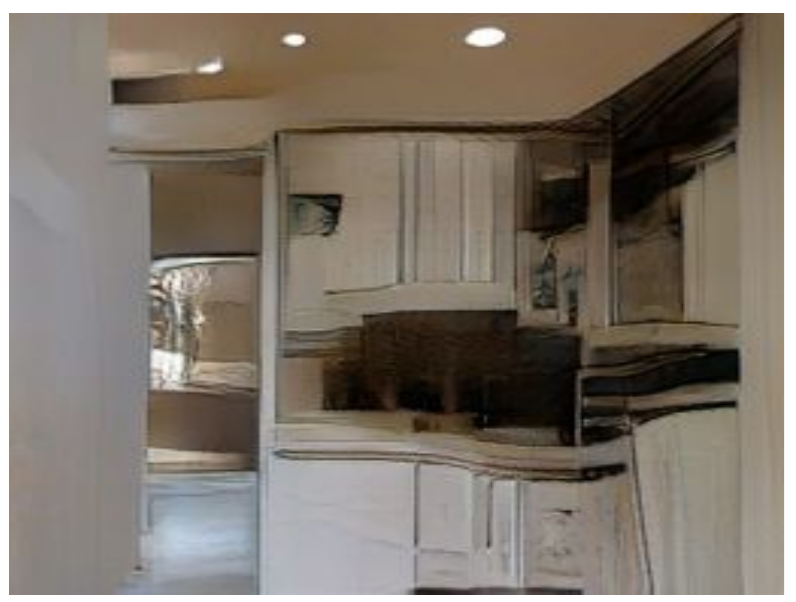} 
    \end{minipage}  & \begin{minipage}{.2\columnwidth}
      \includegraphics[width=\columnwidth]{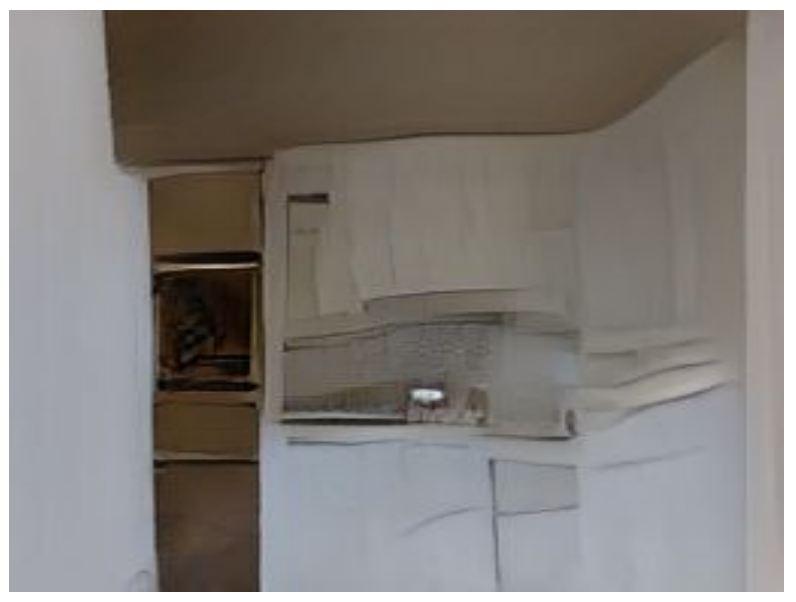}
    \end{minipage} \\
    \begin{minipage}{.2\columnwidth}
     \includegraphics[width=\columnwidth]{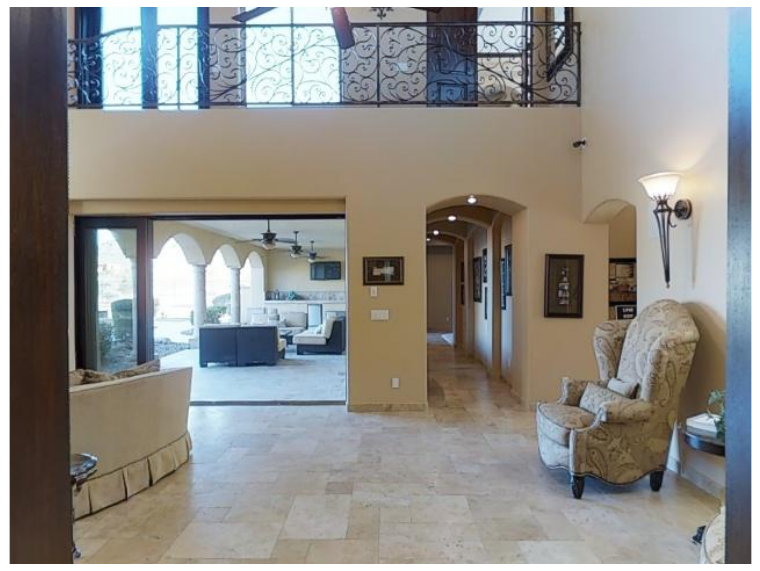} 
    \end{minipage} & \begin{minipage}{.2\columnwidth}
      \includegraphics[width=\columnwidth]{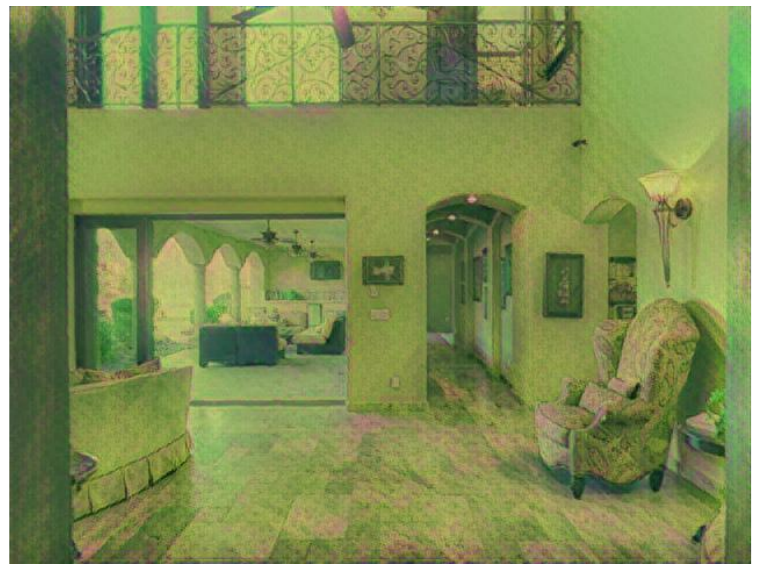} 
    \end{minipage} & \begin{minipage}{.2\columnwidth}
      \includegraphics[width=\columnwidth]{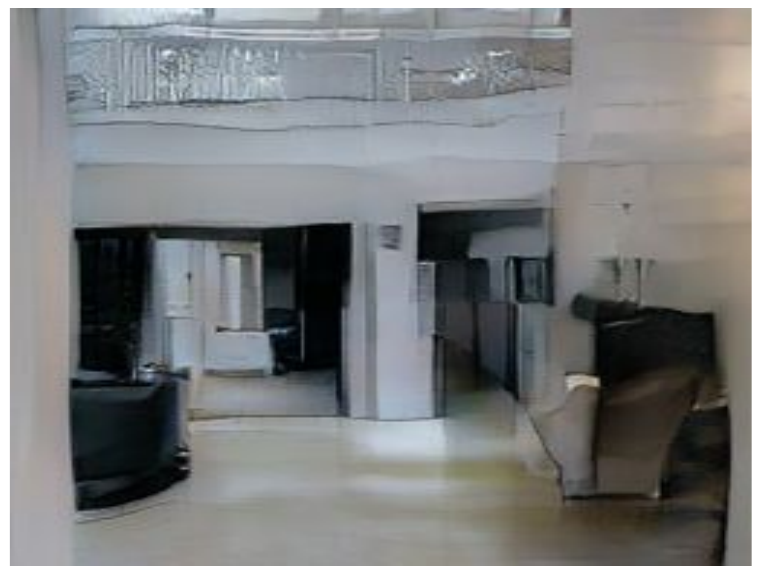} 
    \end{minipage}  & \begin{minipage}{.2\columnwidth}
      \includegraphics[width=\columnwidth]{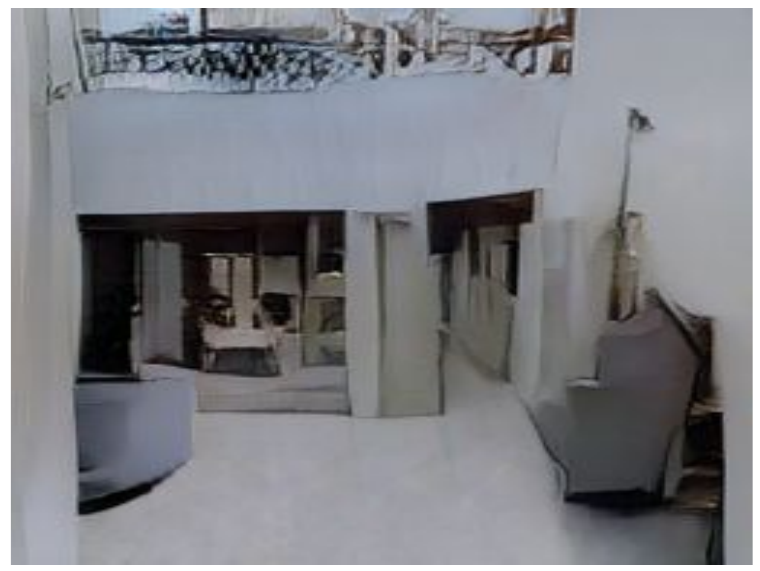}
    \end{minipage} \\
    \begin{minipage}{.2\columnwidth}
     \includegraphics[width=\columnwidth]{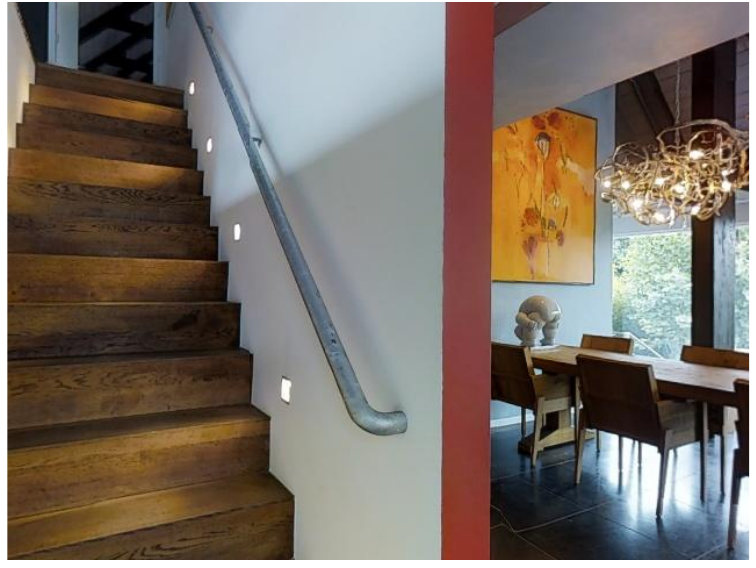} 
    \end{minipage} & \begin{minipage}{.2\columnwidth}
      \includegraphics[width=\columnwidth]{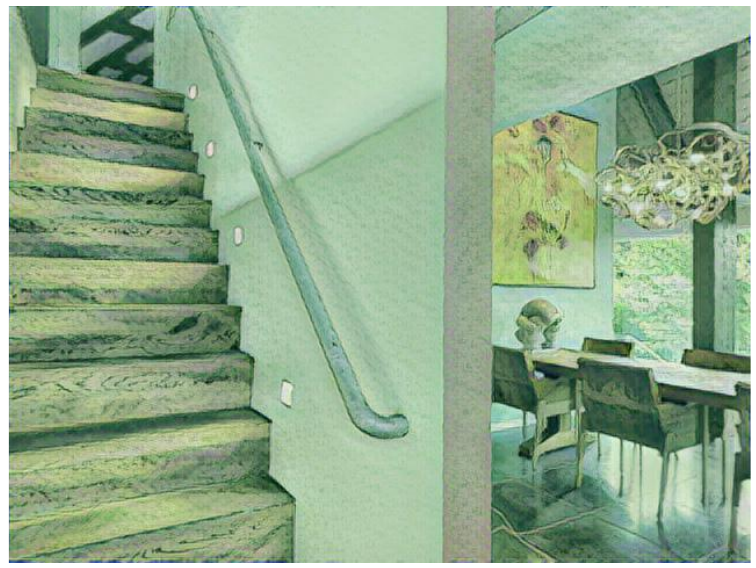} 
    \end{minipage} & \begin{minipage}{.2\columnwidth}
      \includegraphics[width=\columnwidth]{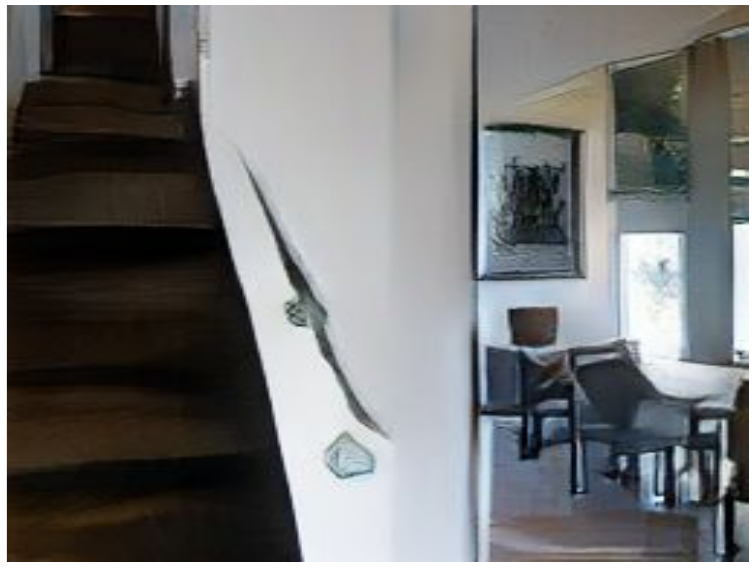} 
    \end{minipage}  & \begin{minipage}{.2\columnwidth}
      \includegraphics[width=\columnwidth]{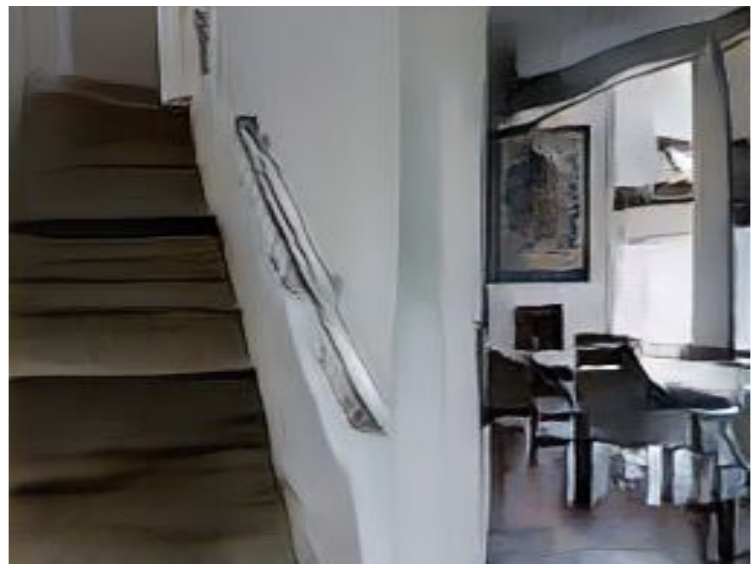}
    \end{minipage} \\
    \begin{minipage}{.2\columnwidth}
     \includegraphics[width=\columnwidth]{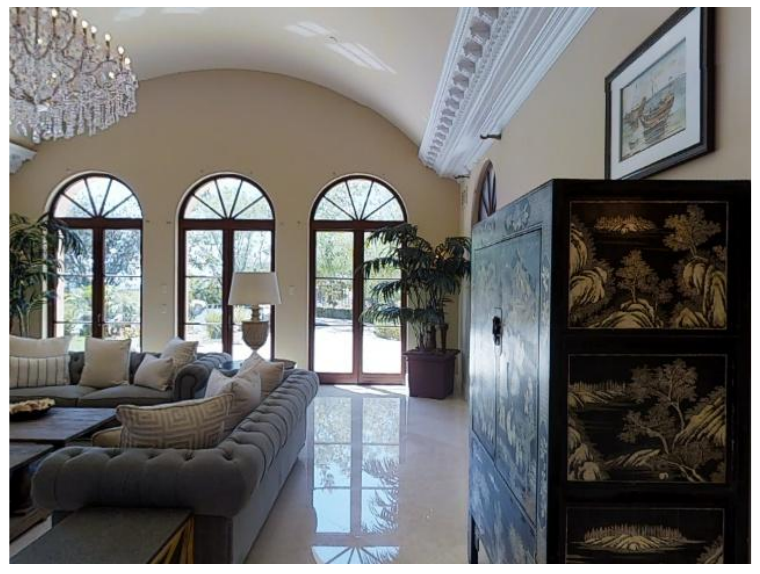} 
    \end{minipage} & \begin{minipage}{.2\columnwidth}
      \includegraphics[width=\columnwidth]{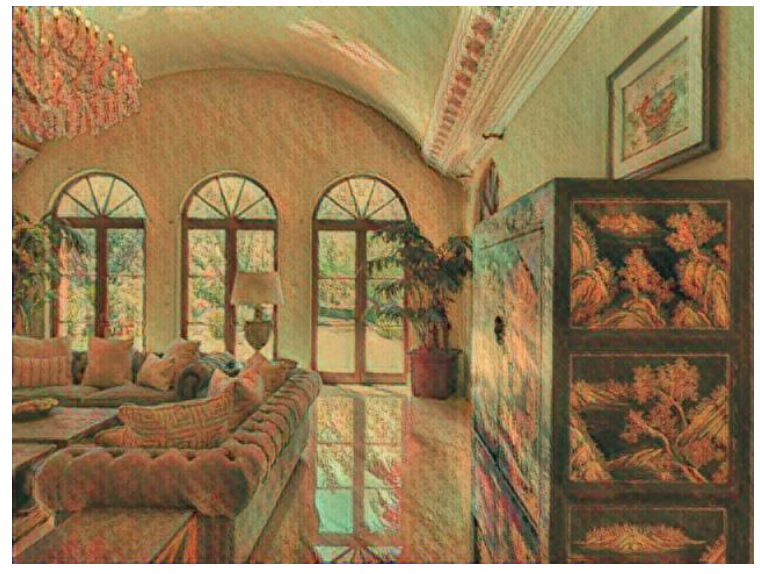} 
    \end{minipage} & \begin{minipage}{.2\columnwidth}
      \includegraphics[width=\columnwidth]{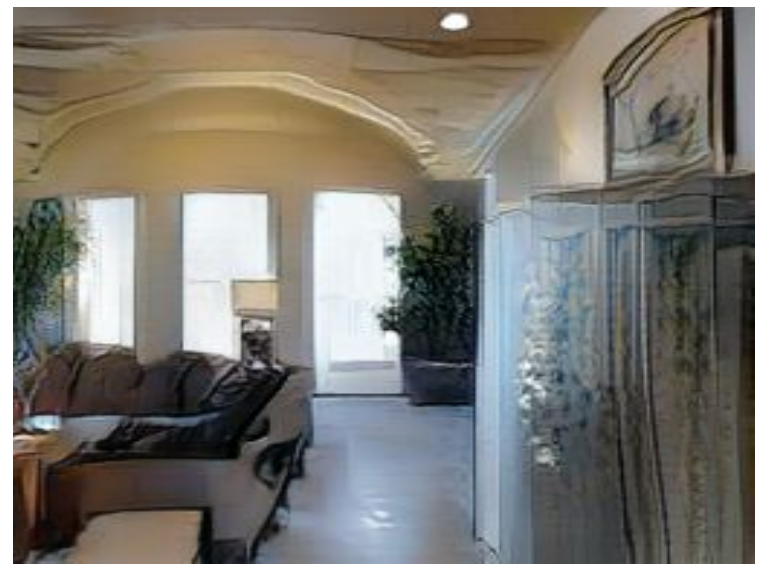} 
    \end{minipage}  & \begin{minipage}{.2\columnwidth}
      \includegraphics[width=\columnwidth]{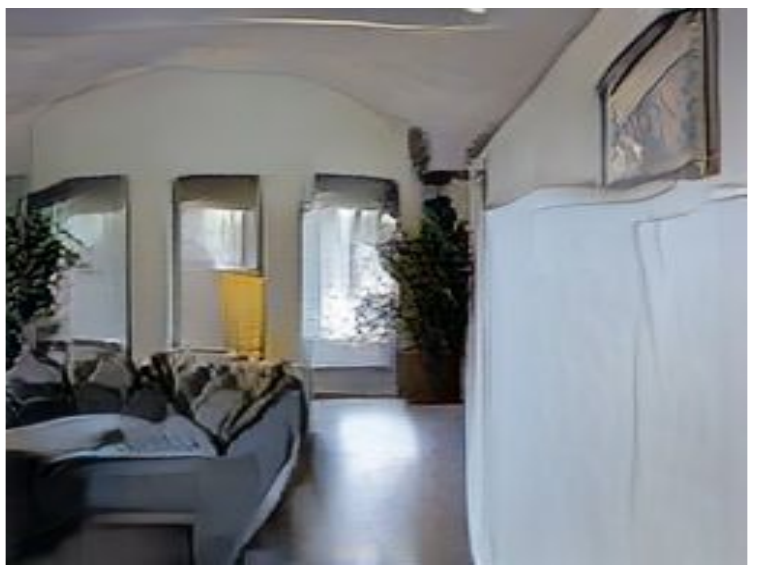}
    \end{minipage} \\
    \hline
    \end{tabular}
    }
    \caption{Qualitative Examples of our edited environments.}
    \vspace{-13pt}
    \label{table7}
\end{table}

\subsection{Test Set Results} \label{sec:test}
We show our method's performance on both the Room-to-Room (R2R) and the multi-lingual Room-Across-Room (RxR) leaderboards. All our agents are tested under the single-run setting, where the agent only navigates once and does not pre-explore the test environment.

On R2R dataset, we first compare our \methodname{} with non-pre-training methods. Specifically, we apply our \methodname{} to EnvDrop-CLIP \cite{shen2021much}, and train the model on $E_{is_1}^m$ with ViT-B/16 features. As shown in Table~\ref{table5}, it outperforms the previous best non-pre-training method (``EnvDrop-CLIP") by 1.6\% in SR and 1.4\% in SPL on test unseen set.  We further adapt our \methodname{} to pre-trained SotA model HAMT \cite{chen2021history}. The model shown in Table~\ref{table5} is an ensemble of models trained on $E_{st}$, $E_{is_1}$ and $E_{is_1}^m$. Our \methodname{} outperforms the HAMT by 3.2\% in SR and 3.9\% in SPL, and achieves the new SotA on the leaderboard.

On the multi-lingual RxR dataset, we first apply our \methodname{} on non-pre-trained SotA model EnvDrop-CLIP \cite{shen2021much}, where we replace the LSTM based instruction encoder with multi-lingual BERT. We train the model with $E_{st}$, and utilize ViT-B/16 to extract visual features. As shown in Table~\ref{table5}, our \methodname{} surpasses the previous best non-pre-training method (``EnvDrop-CLIP") by 5.3\% in nDTW and 8.0\% in sDTW.
We further adapt our \methodname{} to pre-trained SotA model HAMT \cite{chen2021history}. For a fair comparison with HAMT, we use ViT-B/32 features and do not train the visual backbone end-to-end. Ensembling VLN agents trained on three edited environments ($E_{st}$, $E_{is_1}$, $E_{is_1}^m$) outperforms HAMT by 4.7\% in nDTW and 6.6\% in sDTW on the test leaderboard, achieving the new SotA for RxR dataset.

\subsection{Qualitative Analysis for Edited Environments} \label{sec:qual}
We show some examples for our edited environments in Table~\ref{table7}. We could see that the environments generated with the style transfer approach ($E_{st}$) maintain the semantics of the original environments better, with the style being artistic.
The environments generated with the image synthesis approach ($E_{is_1}$ and $E_{is_1}^m$) change object appearance and are more close to real environments. For example, in the last row of $E_{is_1}^m$, the cabinet is masked out during image generation, which brings more diversity in the environments.

\section{Conclusion and Discussion}
In this paper, we present \methodname{}, which augments the Vision-and-Language Navigation training by editing existing environments. Our created environments differ from the original environment in the overall style, object appearance, and object classes, thus can mimic the unseen environments. Our experiments on both Room-to-Room and Room-Across-Room datasets show that training on the edited environments improves the performance in all evaluation metrics compared with both pre-training and non-pre-training methods, and achieves the new SotA on both test leaderboards. Furthermore, we ensemble the VLN agents trained on different edited environments and show that these editing methods are complementary to each other.

\section*{Acknowledgement}
We thank the reviewers, Jaemin Cho, Yi-Lin Sung, Hyounghun Kim, Jie Lei, and Zhenlin Xu for their helpful comments. This work was supported by NSF-CAREER Award 1846185, ARO W911NF2110220, ONR N000141812871, DARPA KAIROS FA8750-19-2-1004, Google Focused Award. The views contained in this article are those of the authors and not of the funding agency.

{\small
\bibliographystyle{ieee_fullname}
\bibliography{egbib}
}

\appendix

\section{Appendix Overview}
In this supplementary, we first describe the implementation details of our editing methods and VLN training in Sec.~\ref{sec:imp} and explain the evaluation metrics we use in Sec.~\ref{sec:eval_metrics}. We show the back translation results for models trained with CLIP-ViT-32 and CLIP-ViT-16 features on different created environments in Sec.~\ref{sec:bt}, and adapt our \methodname{} to another pretrained VLN agent in Sec.~\ref{sec:rec-appendix}. We explore the impact of image preprocessing during feature extraction in Sec.~\ref{sec:rc}. We then explore curriculum learning for training on both the original environments and the created environments in Sec.~\ref{sec:cl}. Furthermore, we show the performance of training with the environments created by masking out different numbers of classes in Sec.~\ref{sec:spade}. Moreover, we include our agent's performance for each different unseen environments in Sec.~\ref{sec:eeval} and analyze the performance variation when adapting different edited environments to different VLN base agents in Sec.~\ref{sec:con}. Lastly, we show qualitative examples from our created environments in Sec.~\ref{sec:example}. Limitations of our work is discussed in Sec.~\ref{sec:limitation}.

\section{Implementation Details} \label{sec:imp}
In the environment creation stage, the style transfer model is directly adopted from \cite{jackson2019style}. We train the image synthesis model for 10 epochs on the training environments in Room-to-Room dataset with the default hyper-parameters from \cite{park2019semantic}. The model is trained to generate new environments given the semantic segmentation of the original environments. In the Vision-and-Language Navigation training, we use the default hyper-parameters from \cite{tan2019learning, hong2020recurrent} when training their models on R2R dataset. On the RxR dataset, we replace the bi-directional LSTM based instruction encoder in \cite{shen2021much} with a cased multi-lingual $\text{BERT}_\text{BASE}$ \cite{devlin2018bert}. The instruction encoder is optimized with AdamW \cite{kingma2014adam} and linear-decayed learning rate (peak at 4e-5). Other parameters are the same as in \cite{shen2021much}. In the back translation stage, the speaker and the agent are trained with the same hyper-parameters as in \cite{tan2019learning}. 

\section{Evaluation Metrics} \label{sec:eval_metrics}
We evaluate our model with six metrics: (1) Success Rate (SR): whether the agent could stop within 3 meters of the target. (2) Success Rate weighted by Path Length (SPL) \cite{anderson2018evaluation}, which penalizes navigation with long paths. (3) Trajectory Length (TL): path length in meters. (4) Navigation Error (NE): the distance between the target and the agent's end point in meters. (5) normalized Dynamic Time Warping (nDTW) \cite{ilharco2019general}: penalizes agents for deviating from the reference path. (6) success rate weighted by normalized Dynamic Time Warping (sDTW) \cite{ilharco2019general}: only consider the nDTW score for successful paths and set the score to 0 for failed paths. SR, SPL are the main metrics for evaluation on R2R dataset, and nDTW, sDTW are the main metrics for RxR dataset. 

\section{Back Translation Results} \label{sec:bt}
We show the performance of training with the original environment and one type of newly-augmented environments (i.e., $E_{st}$, $E_{is_1}$, $E_{is_2}$, $E_{is_1}^m$, $E_{is_2}^m$) on R2R dataset in Table~\ref{table7_appendix}. Back translation with the style-aware speaker is applied in these experiments.
We can see that training with any of our created environments improves the baseline model by a large margin, which is consistent with the performance before back translation. 
Specifically, after back translation, $E_{is_1}$ works the best in SR for ViT-B/32 features, and $E_{is_1}^m$ works the best in SR for ViT-B/16 features, outperforming the baseline model by 4.9\% and 3.1\% respectively. 
Besides, after back translation, training on environments created with image synthesis method gets higher performance than environments created with style transfer approach. Specifically, model trained on environments created with image synthesis method improves 3.5\% with back translation while model trained on environments created with style transfer approach improves 2.1\%\footnote{Performance without back translation is in Table 1 in the main paper}. This indicates models trained with different environments benefit differently from back translation. 

\begin{table*}[]
    \centering
    \begin{tabular}{|c|ccc|cccc|cccc|}
    \hline 
        & \multicolumn{3}{c|}{\textbf{Environment Components}}  & \multicolumn{4}{c|}{\textbf{ViT-B/32}} & 
       \multicolumn{4}{c|}{\textbf{ViT-B/16}}\\  \hline 
       \textbf{Models} & \textbf{Style} & \textbf{Appearance} & \textbf{Object} & \textbf{TL} & \textbf{NE$\downarrow$} & \textbf{SR$\uparrow$} & \textbf{SPL$\uparrow$}   & \textbf{TL} & \textbf{NE$\downarrow$} & \textbf{SR$\uparrow$} & \textbf{SPL$\uparrow$}  \\ \hline
EnvDrop$^*$ \cite{shen2021much} & \xmark & \xmark & \xmark & 16.128 & 4.794 & 55.6 & 49.7 & 18.844 & 4.429 & 57.9 & 50.6 \\ \hline 
$E_{st}$ & \cmark & \xmark & \xmark & 15.912 & 4.335 & 60.2 & \textbf{53.8} & 15.916 & 4.344 & 60.3 & 53.3 \\
$E_{is_1}$ & \xmark & \cmark & \xmark & 17.495 & \textbf{4.325} & \textbf{60.5} & 53.1 & 16.984 & 4.387 & 59.3 & 52.3 \\ 
$E_{is_2}$ & \cmark & \cmark & \xmark  & 17.945 & 4.650 & 58.6 & 51.1 & 18.189 & 4.423 & 59.1 & 51.1 \\ 
$E_{is_1}^m$ & \xmark & \cmark & \cmark & 17.956 & 4.531 & 58.5 & 51.0 & 17.989 & 4.232 & \textbf{60.8} & 54.2 \\
$E_{is_2}^m$ & \cmark & \cmark & \cmark & 17.283 & 4.405 & 58.2 & 51.1 & 16.204 & \textbf{4.181} & 60.4 & \textbf{54.4}  \\
    \hline
    \end{tabular}
    \caption{Performance of training the agent with one kind of our edited environments and back translation. Results are on R2R val-unseen set. ViT-B/32(16) indicate image features extracted with different CLIP-ViT models \cite{radford2021learning}. ``*" indicates reproduced results. \cmark \  indicates the environment component in the new environment is different from the original environment, while \xmark \  indicates the same.
    }
    \label{table7_appendix}
\end{table*}

\begin{table}[]
    \centering
    \begin{tabular}{ccccc}
    \hline 
      \textbf{Models}   &  \textbf{TL} & \textbf{NE$\downarrow$} & \textbf{SR$\uparrow$} & \textbf{SPL$\uparrow$} \\ \hline
      $\circlearrowright$BERT-p365$^*$ \cite{hong2020recurrent} & 12.425 & 4.104 & 62.1 & 56.3 \\
        $E_{st}$-p365 & 12.551 & 4.054 & 62.4 & 56.3 \\
        $E_{is_1}$-p365 & 11.749 & 3.930 & 62.6 & \textbf{57.4} \\ 
        $E_{is_1}^m$-p365 & 11.739 & \textbf{3.917} & \textbf{62.7} & 56.7 \\
    \hline
        $\circlearrowright$BERT-16 \cite{hong2020recurrent} & 11.564 & 4.019 & 61.9 & 57.1   \\
        $E_{st}$-16 & 12.233 & 3.880 & 63.5 & 57.6 \\
        $E_{is_1}$-16 & 11.956 & \textbf{3.808} & \textbf{64.9} & \textbf{58.1} \\ 
        $E_{is_1}^m$-16 & 12.023 & 3.937 & 63.2 & 57.4 \\ \hline 
    \end{tabular}
    \caption{Performance of applying our proposed method to SotA VLN agents on R2R validation unseen set. ``-16" indicates image features extracted with ViT-B/16, ``-p365" indicates image features extracted with ResNet \cite{he2016deep} pre-trained on ImageNet \cite{deng2009imagenet} and finetuned on Place365 \cite{zhou2017places}. ``*" indicates reproduced results.}
    \label{table2_appendix}
\end{table}

\section{Performance on Rec-BERT} \label{sec:rec-appendix}

In this section, we show that our \methodname{} is complementary to other VLN pre-training methods. We enhance the VLN pre-traind model \cite{hong2020recurrent} with our methods and illustrate the improvements on R2R dataset.

The model architecture of \cite{hong2020recurrent} is based on transformer. The weights of their model are initialized with PREVALENT \cite{hao2020towards}, a multi-modal transformer pre-trained on in-domain VLN data. In both works, image features are extracted with the ResNet \cite{he2016deep} which is pre-trained on ImageNet \cite{deng2009imagenet} and fine-tuned on Place365 \cite{zhou2017places}. We first show results using the same image features as their work, and further explore adapting ViT-B/16 features for their model.

As shown in Table~\ref{table2_appendix}, when trained with ``place365" features, augmenting the original environment with $E_{is_1}^m$ could improve the baseline by 0.5\% in SR and 1.1\% in SPL. Augmenting with the other two environments could also improve the baseline by 0.5\% in both SR and SPL. This demonstrates the effectiveness of adapting our method to other strong SotA VLN models.

To use ``ViT-B/16" features, we map the 512 dimension feature to 2048 with a linear layer with dropout.\footnote{Since \cite{hao2020towards} does not provide source code for their in-domain pre-training method, it's hard to start from pre-training a multi-modal transformer with ``ViT-B/16" features and then transfer to \cite{hong2020recurrent} model.} Though this way of adapting features could not fully take advantage of in-domain pre-training \cite{hao2020towards} on ``places365" features, the baseline ``$\circlearrowright$BERT-16" shows competitive performance compared with ``$\circlearrowright$BERT-place365" on both SR and SPL. Augmenting the original environments with $E_{is_1}$ achieves the best performance, improving the baseline by 3.0\% in SR and 1.0\% in SPL. Similar improvements are observed for the other two kinds of new environments (i.e., 1.6\% in SR for $E_{st}$ and 1.3\% in SR for $E_{is_1}^m$), validating the effectiveness of our proposed approach other over SotA methods. 

\section{Image Preprocessing Variants} \label{sec:rc}

In this section, we explore whether image preprocessing will influence the performance when adapting \methodname{} to SotA pre-trained VLN agents $\circlearrowright$BERT \cite{hong2020recurrent}. Our experiments are based on features extracted with ResNet \cite{he2016deep} pre-trained on ImageNet \cite{deng2009imagenet} and fine-tuned on Place365 \cite{zhou2017places}. 

Our image input is of $640\times480\times3$. In the first preprocessing, we resize the image to $256\times256\times3$ and then crop the image at the center to $224\times224\times3$ before normalization, which is consistent with the image preprocessing when evaluating on ImageNet classification task.
We then directly adopt the features from the last pooling layer of the ResNet. We use this preprocessing in our main paper. In the second preprocessing, instead of transforming the image size to match the required ResNet input size, we follow \cite{anderson2018vision} to change the last pooling layer in ResNet to global pooling, and does not further do any resizing on input image except normalization. This preprocessing is also used in previous VLN works.

We show the results for two preprocessing methods in Table~\ref{table8_appendix}. We can see that removing the image transformation in the preprocessing significantly improves the baseline performance ($\circlearrowright$BERT-1 vs. $\circlearrowright$BERT-2). This is due to the baseline model is pre-trained on ``place365" features extracted by \cite{anderson2018vision}, which does not use image resize and crop. We also observe that after training on both the original environment and our edited environment, the performance gap decreases ($E_{is_1}$-1 vs. $E_{is_1}$-2).

\begin{table}[]
    \centering
    \begin{tabular}{ccccc}
    \hline 
      \textbf{Models}   &  \textbf{TL} & \textbf{NE$\downarrow$} & \textbf{SR$\uparrow$} & \textbf{SPL$\uparrow$} \\ \hline
      $\circlearrowright$BERT-1$^*$ \cite{hong2020recurrent} & 12.421 & 4.294 & 59.5 & 53.6 \\
        $E_{st}$-1 & 12.551 & 4.054 & 62.4 & 56.3 \\
        $E_{is_1}$-1 & 11.749 & 3.930 & 62.6 & \textbf{57.4} \\ 
        $E_{is_1}^m$-1 & 11.739 & \textbf{3.917} & \textbf{62.7} & 56.7 \\
    \hline
        $\circlearrowright$BERT-2$^*$ \cite{hong2020recurrent} & 11.327 & 4.028 & 61.8 & 56.3   \\
        $E_{st}$-2 & 11.760 & 3.921 & 62.8 & 56.6 \\
        $E_{is_1}$-2 &  12.319 & \textbf{3.884} & \textbf{63.2} & \textbf{56.8} \\ 
        $E_{is_1}^m$-2 & 12.034 & 4.016 & 62.3 & 56.3 \\ \hline 
    \end{tabular}
    \caption{Performance of applying our proposed method to SotA VLN agents on R2R validation unseen set. ``-1"/``-2" indicates using the first/second way of preprocess, ``*" indicates results with our extracted ``place365" features.}
    \label{table8_appendix}
\end{table}

\begin{table}[]
    \centering
    \begin{tabular}{ccccc}
    \hline 
      \textbf{Models}   &  \textbf{TL} & \textbf{NE$\downarrow$} & \textbf{SR$\uparrow$} & \textbf{SPL$\uparrow$} \\ \hline
      EnvDrop$^*$ \cite{shen2021much} & 15.86 & 4.73 & 55.1 & 48.8 \\
        $E_{st}$ & 16.59 & 4.69 & 58.2 & 51.5 \\
        $E_o$ + $E_{st}$ + CL & 20.98 & 5.14 & 49.8 & 43.3 \\
        $E_{st}$ + $E_o$ + CL &  16.08 & 4.63 & 56.8 & 50.5 \\
        $E_{is_1}$ & 17.69 & 4.76 & 56.4 & 48.9 \\
        $E_o$ + $E_{is_1}$ + CL & 26.43 & 5.60 & 47.3 & 39.1 \\ 
        $E_{is_1}$ + $E_o$ + CL & 17.27 & 4.95 & 55.3 & 48.0 \\ 
        $E_{is_1}^m$ & 14.46 & 4.67 & 57.3 & 51.1 \\
        $E_o$ + $E_{is_1}^m$ + CL & 26.54 & 5.77 & 48.4 & 38.9 \\ 
        $E_{is_1}^m$ + $E_o$ + CL &  17.31 & 4.83 & 55.5 & 48.2 \\
    \hline
    $E_{st}$ + $E_{st_2}$ + $E_o$ + CL & 15.85 & 4.67 & 56.2 & 49.8 \\ 
    $E_{is_1}$ + $E_{st}$ + $E_o$ + CL & 17.47 &  4.83 & 56.2 & 49.1 \\ \hline
    \end{tabular}
    \caption{Performance of training the agent on the original environment and edited environment in different steps in curriculum learning on R2R validation unseen set. ``-CL" indicates using curriculum learning, ``*" indicates reproduced results.}
    \label{table9_appendix}
\end{table}

\section{Curriculum Learning} \label{sec:cl}
In the main paper, we train the agent on both the original environment and the edited environment in the same batch. 
This section explores using curriculum learning to train the agent on the original environment and edited environment in different steps. Specifically, we train the models with two steps. In the first step, the model is trained on one of the newly created environments. In the second step, the model is trained on the original environments. Both steps contain 50,000 iterations with a batch size of 64. We use image features extracted with ViT-B/16 in this experiment. For simplicity, back translation is not applied in this experiment.

We notice that curriculum learning usually learns from easy samples first and gradually switch to hard samples. In our case, intuitively, easy samples should be the original environment and hard samples are our created environments. However, the performance of models first trained on original environment and then trained on newly created environments (as shown in Table~\ref{table9_appendix}) are much worse than training in the reverse way (e.g., ``$E_o + E_{st}$ + CL" vs. ``$E_{st}$"). We attribute this to that the data in original environments are of higher quality. 

\begin{table}[]
    \centering
    \begin{tabular}{ccccc}
    \hline 
      \textbf{Models}   &  \textbf{TL} & \textbf{NE$\downarrow$} & \textbf{SR$\uparrow$} & \textbf{SPL$\uparrow$} \\ \hline
        $E_{is_1}^{m_1}$-16 & 14.46 & 4.67 & 57.3 & 51.1 \\
        $E_{is_1}^{m_2}$-16 & 17.01 & 4.86 & 55.6 & 48.7 \\ 
        $E_{is_1}^{m_3}$-16 & 17.05 & 4.94 & 54.6 & 48.4 \\ 
        $E_{is_1}^{m_4}$-16 & 16.54 & 4.95 & 54.2 & 47.7 \\
    \hline
    \end{tabular}
    \caption{Performance of training on the environments created by masking out different number of objects on R2R validation unseen set. ``-16" indicates image features extracted with ViT-B/16.}
    \label{table10_appendix}
\end{table}

\begin{table*}[h]
    \centering
    \begin{tabular}{c>{\centering\arraybackslash}c|>{\centering\arraybackslash}p{0.1\columnwidth}>{\centering\arraybackslash}p{0.1\columnwidth}|>{\centering\arraybackslash}p{0.1\columnwidth}>{\centering\arraybackslash}p{0.1\columnwidth}|>{\centering\arraybackslash}p{0.1\columnwidth}>{\centering\arraybackslash}p{0.1\columnwidth}|>{\centering\arraybackslash}p{0.1\columnwidth}>{\centering\arraybackslash}p{0.1\columnwidth}}
\hline 
\multicolumn{1}{c}{\textbf{ID}} &
\multicolumn{1}{c}{\textbf{Environment}} &  \multicolumn{2}{c}{\textbf{EnvDrop-16$^*$ \cite{shen2021much}}} & \multicolumn{2}{c}{$E_{st}$-16}  & \multicolumn{2}{c}{$E_{is_1}$-16}  & 
\multicolumn{2}{c}{$E_{is_1}^m$-16}  \\
    \hline 
   &  &\textbf{SR} &\textbf{SPL} & \textbf{SR} & \textbf{SPL} & \textbf{SR} &\textbf{SPL} & \textbf{SR} & \textbf{SPL} \\
   1 & \begin{minipage}{.1\columnwidth}
      \includegraphics[width=\columnwidth]{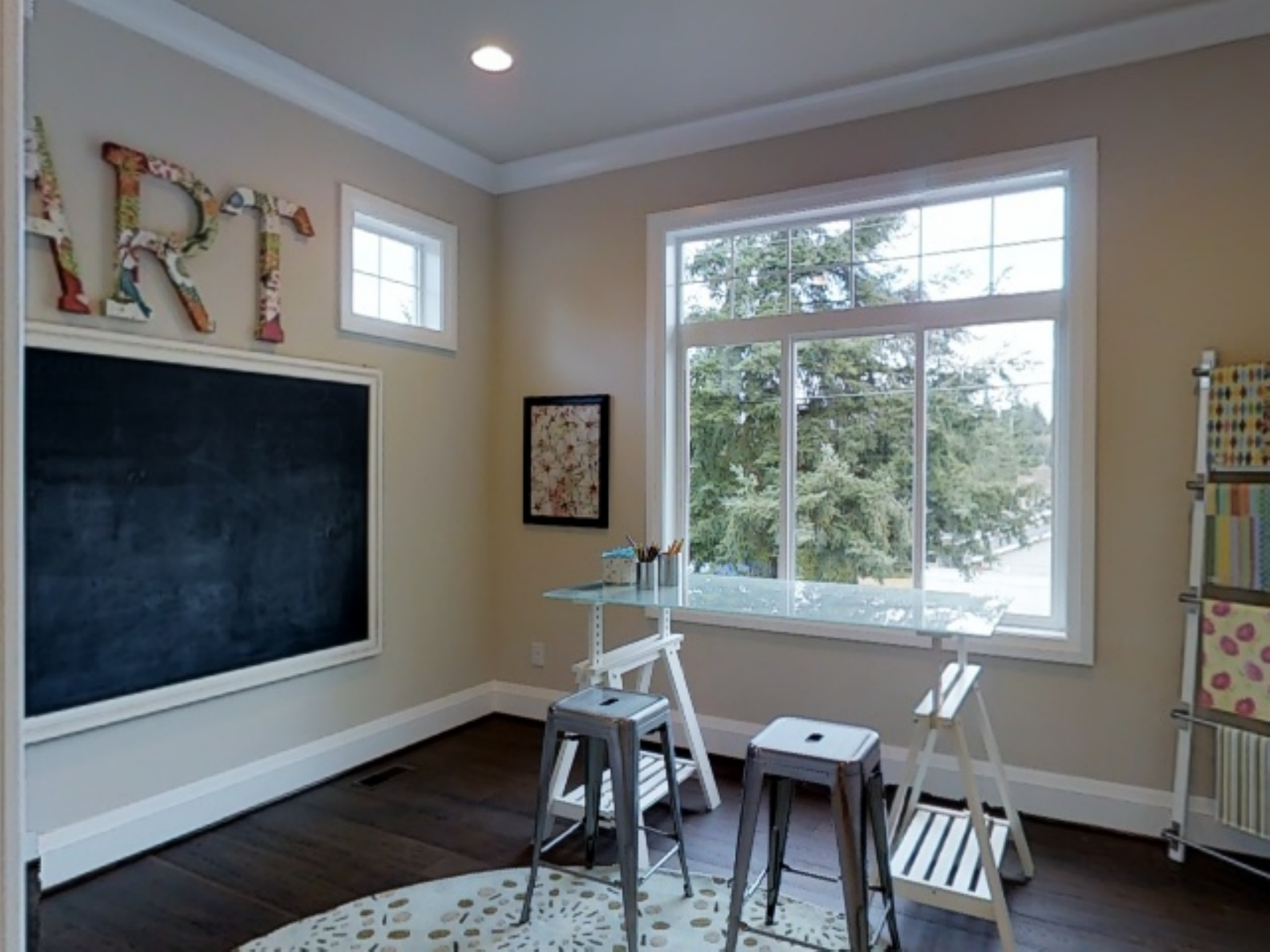} 
    \end{minipage} & 59.3 & 50.4 & 59.7 & 52.3 & 63.3 & 52.1 & 51.3 & 52.1  \\ 
  2 & \begin{minipage}{.1\columnwidth}
      \includegraphics[width=\columnwidth]{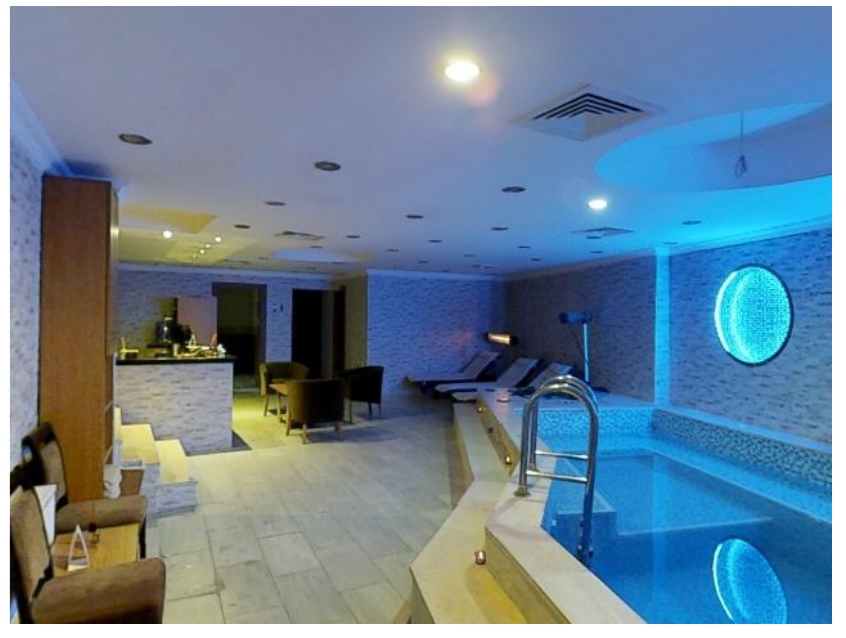} 
    \end{minipage} & 50.4 & 42.8 & 51.1 & 43.8 & 49.7 & 39.7 & 52.5 & 47.1  \\ 
3 & \begin{minipage}{.1\columnwidth}
      \includegraphics[width=\columnwidth]{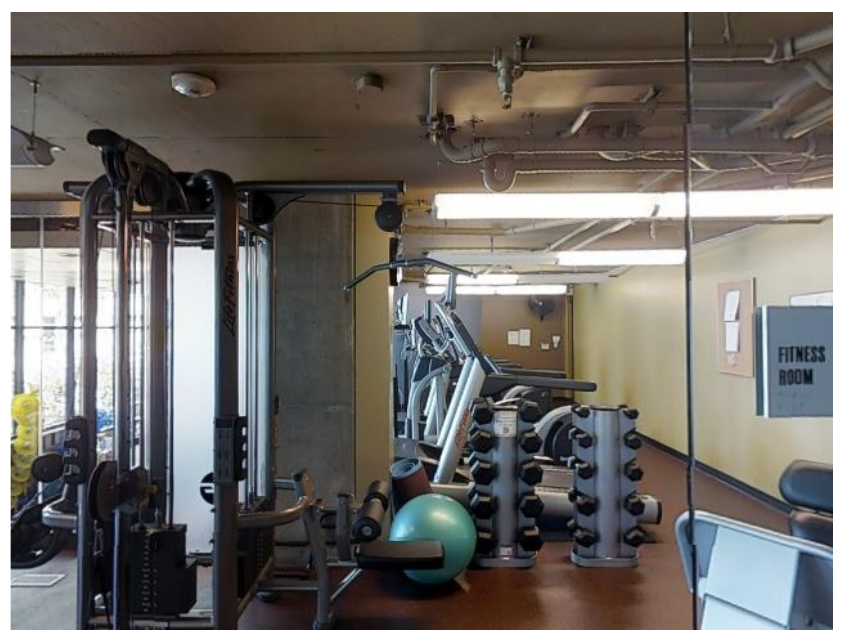} 
    \end{minipage} & 77.8 & 59.8 & 75.6 & 61.1 & 77.8 & 59.9 & 64.4 & 51.5 \\ 
4 & \begin{minipage}{.1\columnwidth}
      \includegraphics[width=\columnwidth]{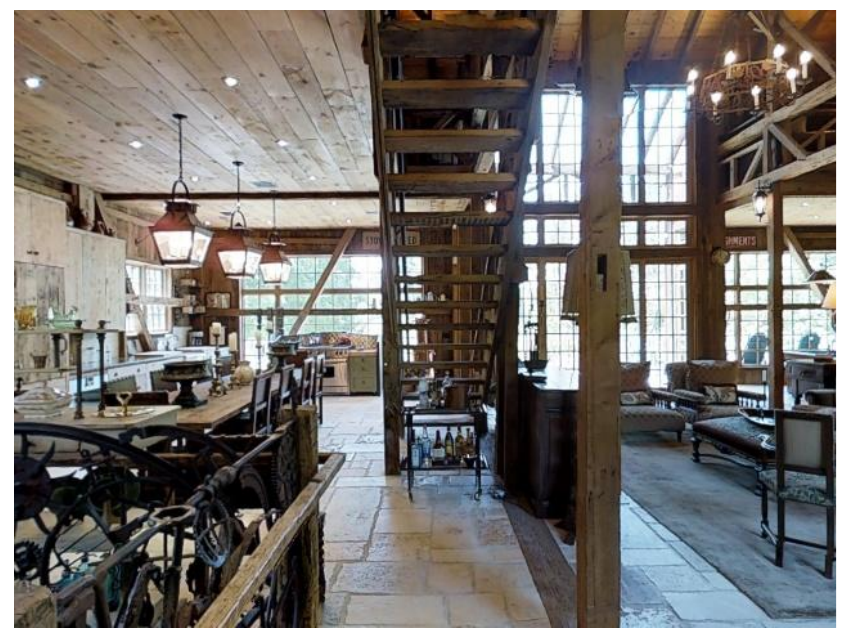} 
    \end{minipage} & 56.8 & 48.7 & 59.9 & 52.0 & 59.4 & 50.0 & 63.5 & 55.4 \\ 
5 & \begin{minipage}{.1\columnwidth}
      \includegraphics[width=\columnwidth]{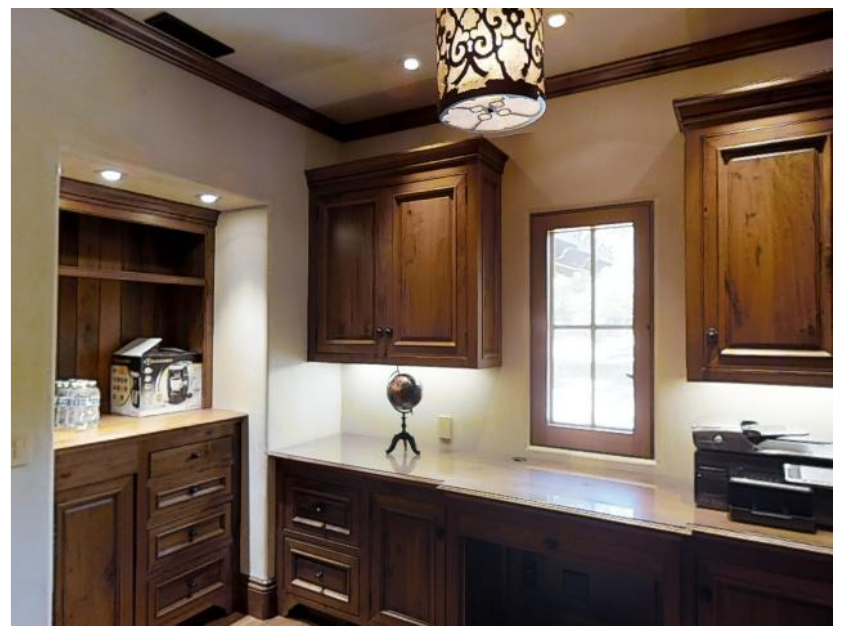} 
    \end{minipage} & 42.7 & 37.1 & 57.7 & 50.9 & 48.3 & 42.4 & 52.7 & 47.0 \\ 
6 & \begin{minipage}{.1\columnwidth}
      \includegraphics[width=\columnwidth]{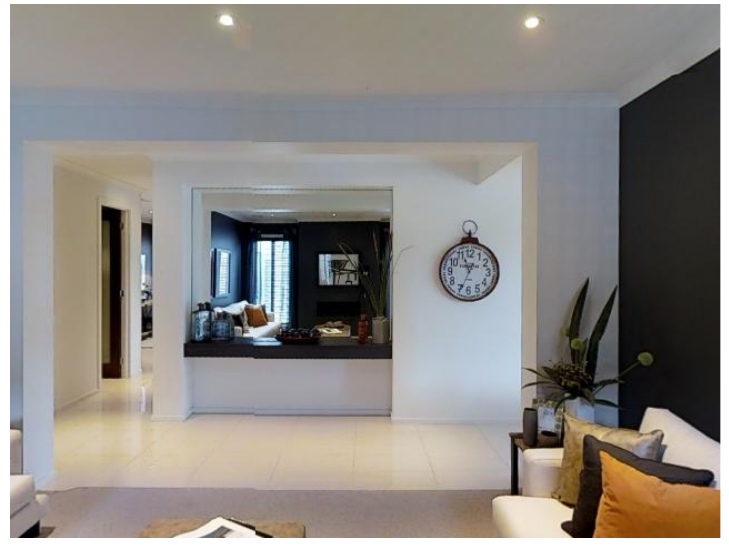} 
    \end{minipage}  & 49.0 & 45.3 & 50.7 & 46.6 & 50.7 & 43.0 & 49.3 & 45.5 \\ 
7 & \begin{minipage}{.1\columnwidth}
      \includegraphics[width=\columnwidth]{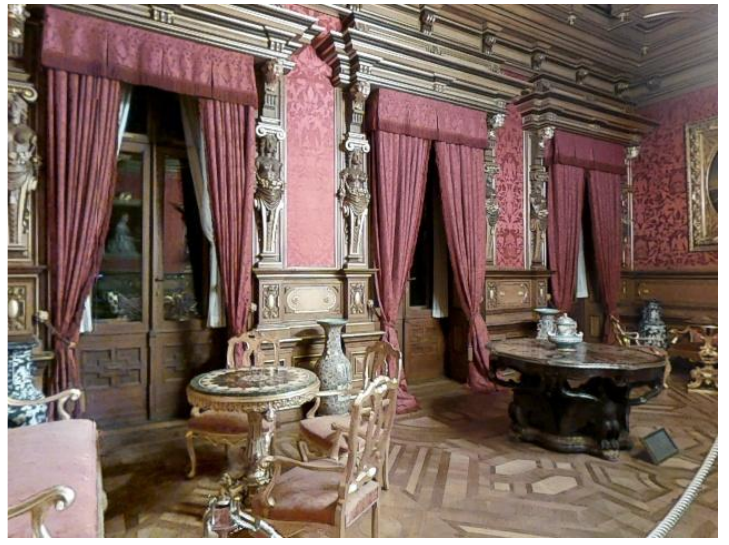} 
    \end{minipage}  & 56.1 & 53.6 & 50.6 & 46.1 & 52.8 & 49.2 & 53.3 & 50.2  \\ 
8 & \begin{minipage}{.1\columnwidth}
      \includegraphics[width=\columnwidth]{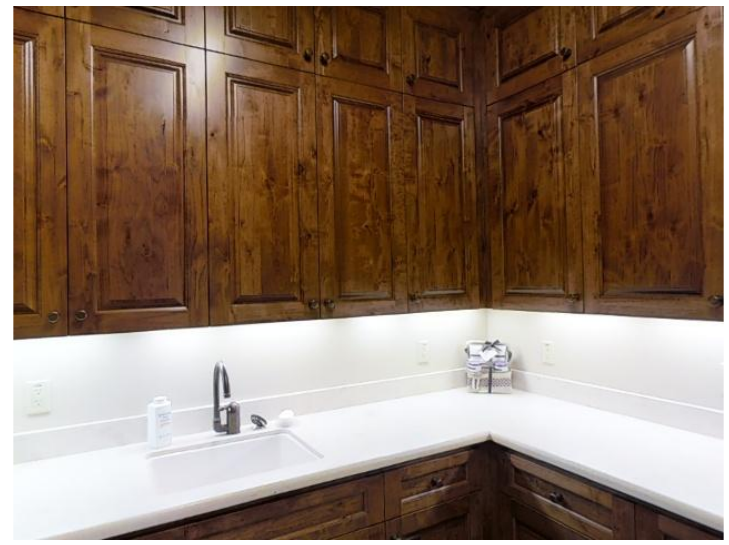}  
    \end{minipage} & 49.5 & 45.3 & 47.0 & 42.8 & 51.6 & 47.6 & 48.0 & 43.8 \\ 
9 & \begin{minipage}{.1\columnwidth}
      \includegraphics[width=\columnwidth]{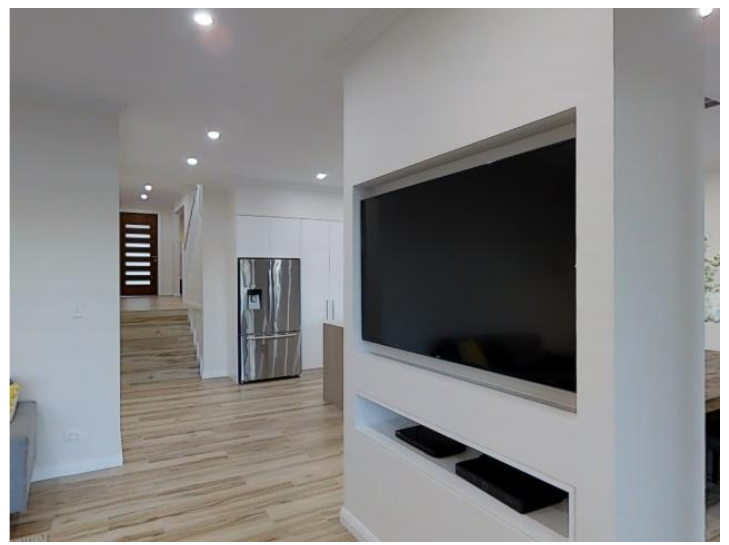}  
    \end{minipage} & 69.1 & 60.1 & 71.4 & 62.1 & 67.0 & 58.5 & 66.3 & 59.9 \\ 
10 & \begin{minipage}{.1\columnwidth}
      \includegraphics[width=\columnwidth]{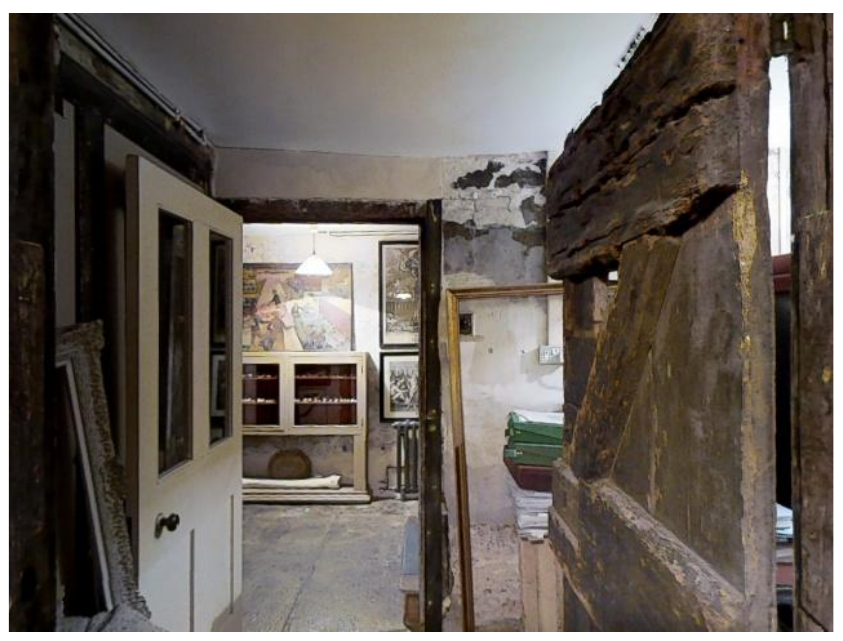}  
    \end{minipage} & 60.7 & 55.5 & 68.0 & 60.1 & 60.3 & 54.2 & 65.0 & 58.2  \\ 
11 & \begin{minipage}{.1\columnwidth}
      \includegraphics[width=\columnwidth]{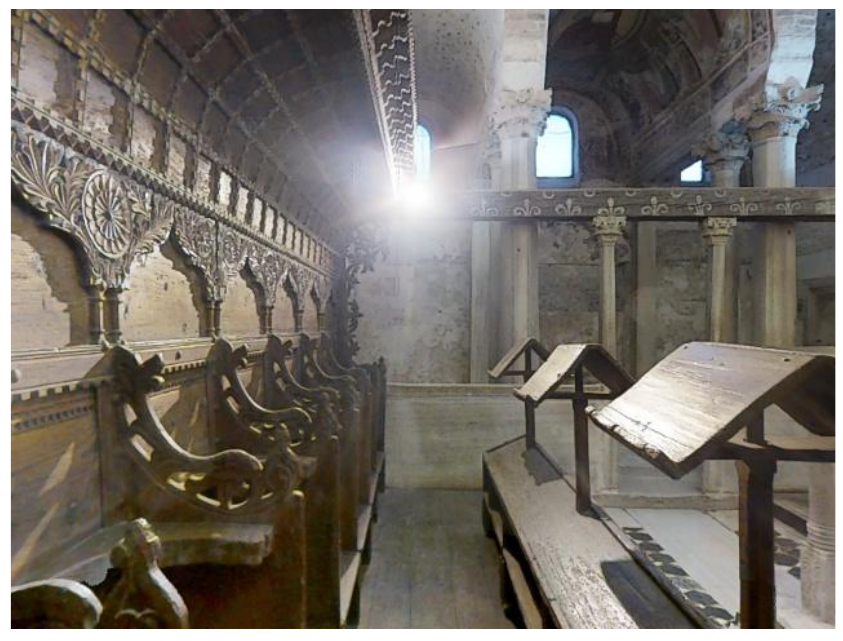}  
    \end{minipage}  & 16.7 & 14.4 & 33.3 & 27.2 & 16.7 & 14.2 &   55.6 & 46.6 \\ 
    \hline
    \end{tabular}
    \caption{The results of our \methodname{} on different environments in validation unseen set. ``-16" indicates image features extracted with ViT-B/16. }
    \label{table11_appendix}
\end{table*}

\begin{table*}[h]
    \centering
    \begin{tabular}{c>{\centering\arraybackslash}c|>{\centering\arraybackslash}p{0.1\columnwidth}>{\centering\arraybackslash}p{0.1\columnwidth}|>{\centering\arraybackslash}p{0.1\columnwidth}>{\centering\arraybackslash}p{0.1\columnwidth}|>{\centering\arraybackslash}p{0.1\columnwidth}>{\centering\arraybackslash}p{0.1\columnwidth}}
\hline 
\multicolumn{1}{c}{\textbf{ID}} &
\multicolumn{1}{c}{\textbf{Environment}} &  \multicolumn{2}{c}{\textbf{EnvDrop-16$^*$ \cite{shen2021much}}} & \multicolumn{2}{c}{\textbf{EnvDrop+BT}}  & \multicolumn{2}{c}{\textbf{$\circlearrowright$BERT-p365$^*$} \cite{hong2020recurrent}}    \\
    \hline 
   &  &\textbf{SR} &\textbf{SPL} & \textbf{SR} & \textbf{SPL} & \textbf{SR} &\textbf{SPL} \\
   1 & \begin{minipage}{.1\columnwidth}
      \includegraphics[width=\columnwidth]{figures/TbHJrupSAjP.pdf} 
    \end{minipage} & 59.3 & 50.4 & 61.3	& 51 & 67.3 & 62.2 \\ 
  2 & \begin{minipage}{.1\columnwidth}
      \includegraphics[width=\columnwidth]{figures/x8F5xyUWy9e.pdf} 
    \end{minipage} & 50.4 & 42.8 & 61 &	50.3 &	58.9& 53.9  \\ 
3 & \begin{minipage}{.1\columnwidth}
      \includegraphics[width=\columnwidth]{figures/8194nk5LbLH.pdf} 
    \end{minipage} & 77.8 & 59.8 & 71.1&	50.6&	77.8&	66.1 \\ 
4 & \begin{minipage}{.1\columnwidth}
      \includegraphics[width=\columnwidth]{figures/EU6Fwq7SyZv.pdf} 
    \end{minipage} & 56.8 & 48.7 & 60.9&	52&	63	&56.8\\ 
5 & \begin{minipage}{.1\columnwidth}
      \includegraphics[width=\columnwidth]{figures/2azQ1b91cZZ.pdf} 
    \end{minipage} & 42.7 & 37.1 & 51.7&	43.7&	55.3&	50.6 \\ 
6 & \begin{minipage}{.1\columnwidth}
      \includegraphics[width=\columnwidth]{figures/zsNo4HB9uLZ.pdf} 
    \end{minipage}  & 49.0 & 45.3 & 55.3&	49.4 &	50.7 &	47.8\\ 
7 & \begin{minipage}{.1\columnwidth}
      \includegraphics[width=\columnwidth]{figures/Z6MFQCViBuw.pdf} 
    \end{minipage}  & 56.1 & 53.6 & 50.6&	49	& 56.1&	53.5  \\ 
8 & \begin{minipage}{.1\columnwidth}
      \includegraphics[width=\columnwidth]{figures/QUCTc6BB5sX.pdf}  
    \end{minipage} & 49.5 & 45.3 & 55.2&	50.4&	63.8&	59.2 \\ 
9 & \begin{minipage}{.1\columnwidth}
      \includegraphics[width=\columnwidth]{figures/X7HyMhZNoso.pdf}  
    \end{minipage} & 69.1 & 60.1 & 66.3 &	57.9&	71.1&	77.9 \\ 
10 & \begin{minipage}{.1\columnwidth}
      \includegraphics[width=\columnwidth]{figures/oLBMNvg9in8.pdf}  
    \end{minipage} & 60.7 & 55.5 & 58.3&	53.5&	65.3&	60.6  \\ 
11 & \begin{minipage}{.1\columnwidth}
      \includegraphics[width=\columnwidth]{figures/pLe4wQe7qrG.pdf}  
    \end{minipage}  & 16.7 & 14.4 & 22.2&	18.9&	61.1&	57.3 \\ 
    \hline
    \end{tabular}
    \caption{The results of different VLN baseline models on different environments in validation unseen set. ``-16" indicates image features extracted with ViT-B/16.}
    \label{table13_appendix}
\end{table*}

Besides, as shown in Table~\ref{table9_appendix}, the model trained with curriculum learning gets lower performance compared with directly training the agent on both the original environment and the edited environment in the same batch. This indicates that more advanced curriculum learning (e.g., a better-designed approach to rank the difficulty of the samples) is needed to get higher performance. 

We further explore using curriculum learning to train the agent on multiple edited environments and the original environment. Specifically, we train the models with three steps. In the first two steps, the agent is trained on two different edited environments. In the last step, the agent is trained on the original environments. All three steps contain 50,000 iterations with a batch size of 64. For simplicity, back translation is not applied in this experiment as well. 

As shown in Table~\ref{table9_appendix}, we first create a new environment $E_{st_2}$ with style transfer approach. The style is sampled from a multivariate normal distribution. We train the agent on $E_{st}$ and $E_{st_2}$ in the first two steps sequentially in curriculum learning. We observe that the performance (``$E_{st}$ + $E_{st_2}$ + $E_o$ + CL") is lower than only training on $E_{st}$. Similar results are observed for training on environments $E_{is_1}$ and $E_{st}$ in the first two steps in curriculum learning (``$E_{is_1}$ + $E_{st}$ + $E_o$ + CL").
We further experiment with adaptive curriculum learning (aCL)
to learn from easy examples to hard examples. The difficulty is measured with path generation probability and variance. We use Bayesian optimization to find the best hyperparameter for aCL based on the agent's performance of the first 10k iterations. The best model achieves 57.6\%, which is still lower than $E_{st}$ only. We believe how to combine the three editing environments during training to get better performance is a non-trivial problem and requires further investigation in future work.

\begin{table*}[h]
    \centering
    \begin{tabular}{ccccc}
\hline 
\multicolumn{1}{c}{\textbf{ID}} &
\multicolumn{1}{c}{\textbf{Original}} &
\multicolumn{1}{c}{$E_{st}$} & 
\multicolumn{1}{c}{$E_{is_1}$} & 
\multicolumn{1}{c}{$E_{is_1}^m$} \\ \hline 
1 & \begin{minipage}{.4\columnwidth}
     \includegraphics[width=\columnwidth]{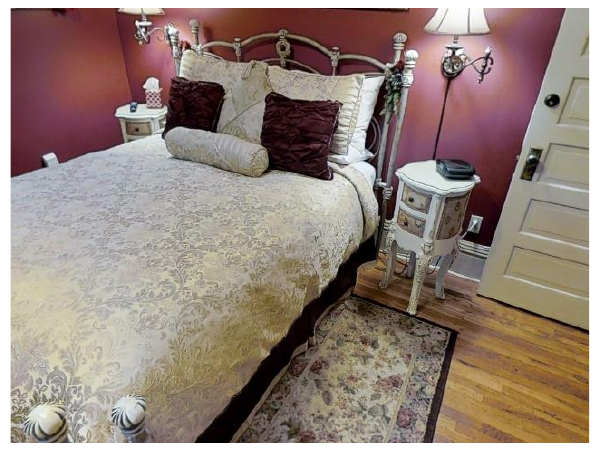} 
    \end{minipage} & \begin{minipage}{.4\columnwidth}
      \includegraphics[width=\columnwidth]{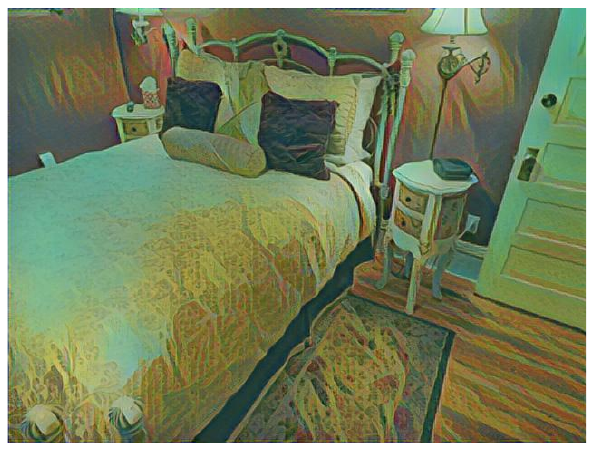} 
    \end{minipage} & \begin{minipage}{.4\columnwidth}
      \includegraphics[width=\columnwidth]{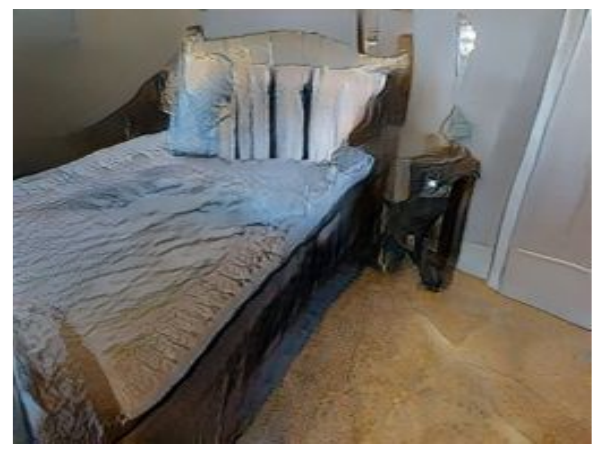} 
    \end{minipage}  & \begin{minipage}{.4\columnwidth}
      \includegraphics[width=\columnwidth]{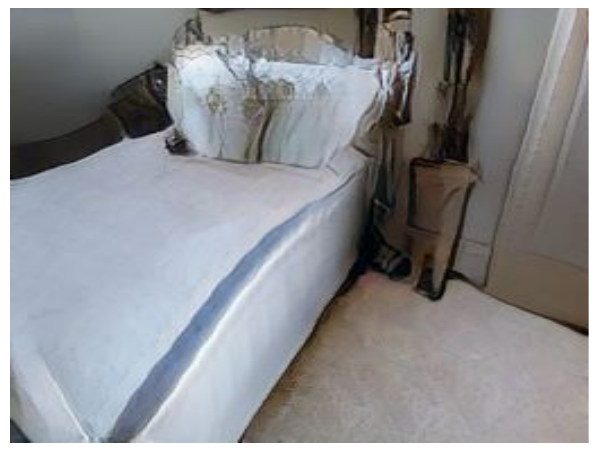} 
    \end{minipage} \\
 2&   \begin{minipage}{.4\columnwidth}
      \includegraphics[width=\columnwidth]{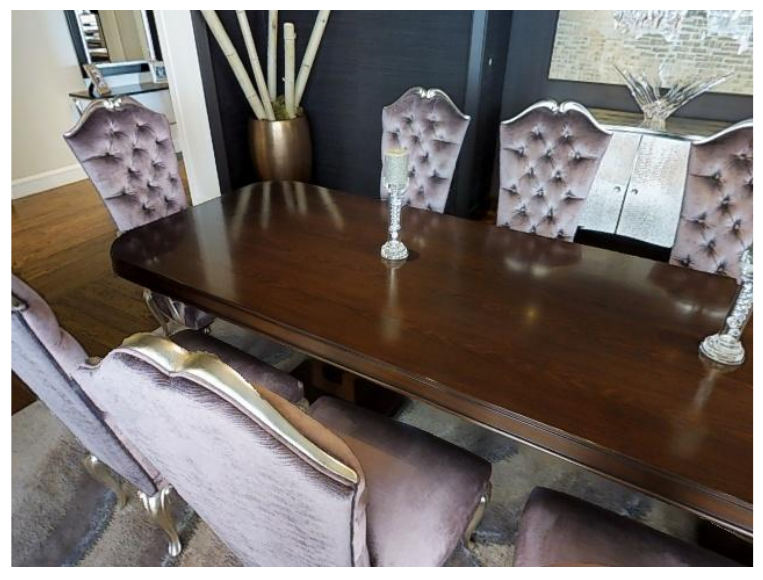} 
    \end{minipage} & \begin{minipage}{.4\columnwidth}
      \includegraphics[width=\columnwidth]{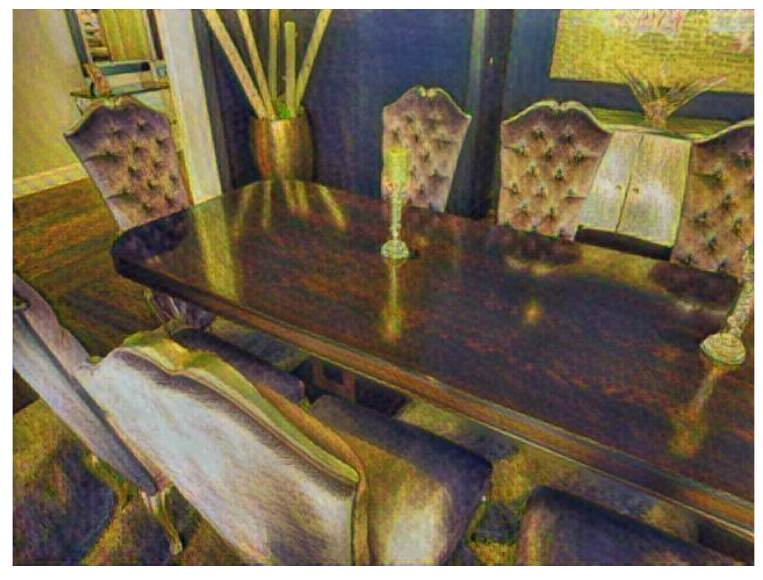} 
    \end{minipage} & \begin{minipage}{.4\columnwidth}
      \includegraphics[width=\columnwidth]{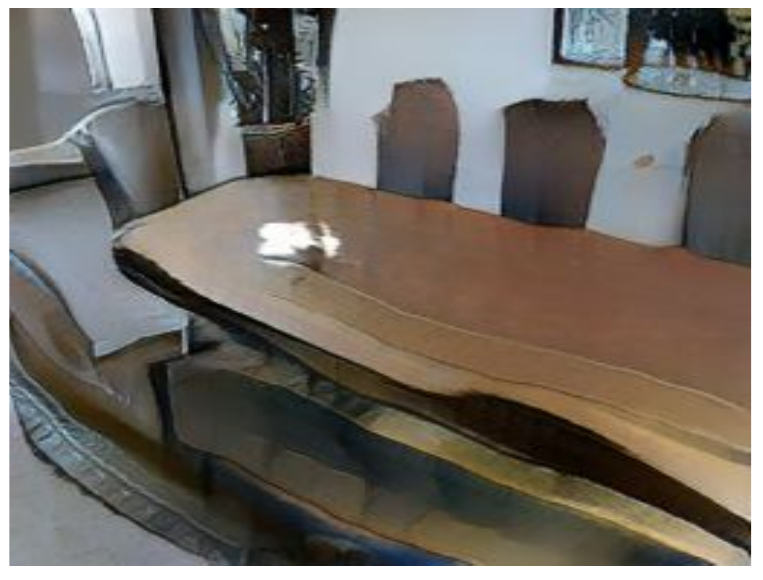} 
    \end{minipage}  & \begin{minipage}{.4\columnwidth}
      \includegraphics[width=\columnwidth]{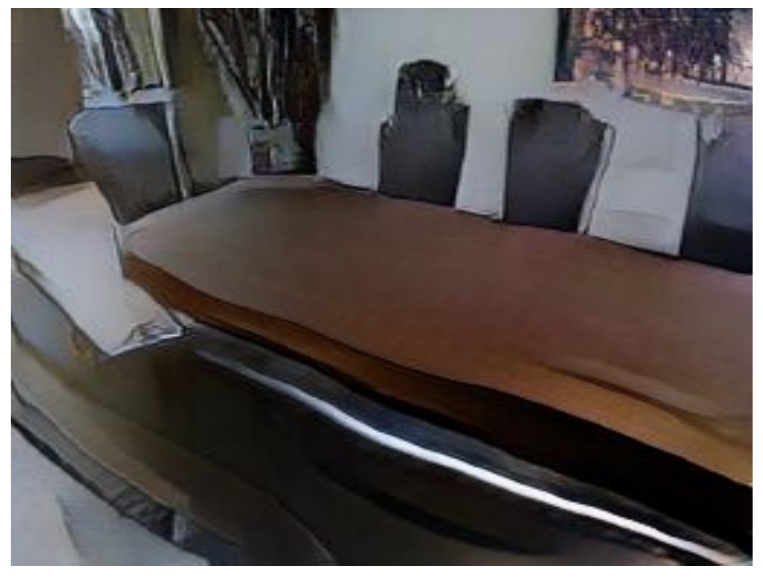} 
    \end{minipage} \\
 3&   \begin{minipage}{.4\columnwidth}
      \includegraphics[width=\columnwidth]{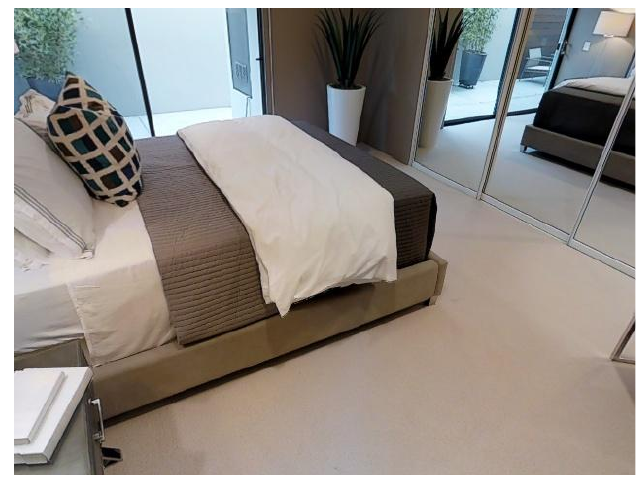} 
    \end{minipage} & \begin{minipage}{.4\columnwidth}
      \includegraphics[width=\columnwidth]{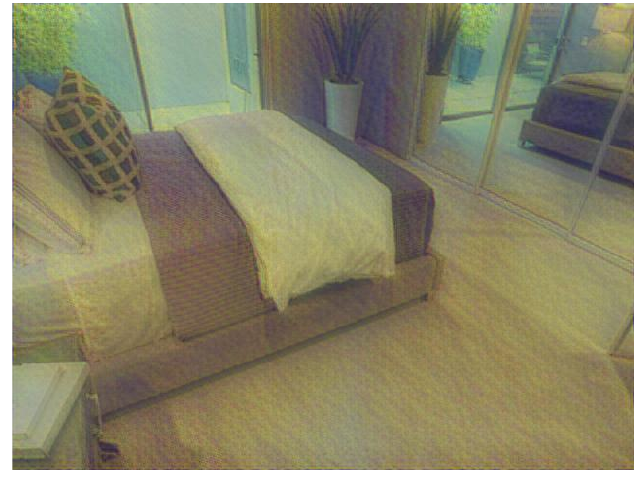} 
    \end{minipage} & \begin{minipage}{.4\columnwidth}
      \includegraphics[width=\columnwidth]{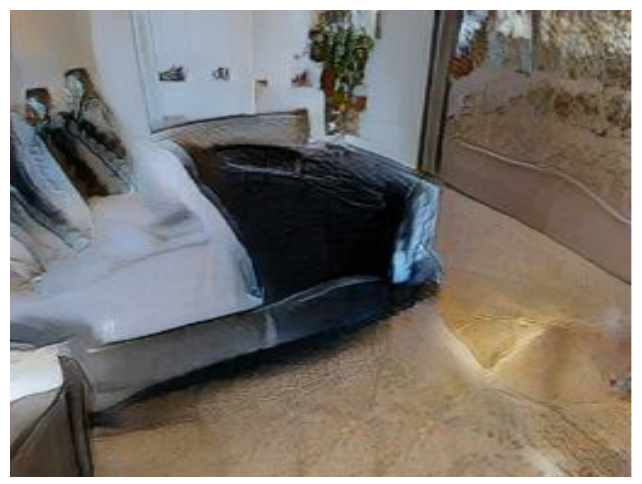} 
    \end{minipage}  & \begin{minipage}{.4\columnwidth}
      \includegraphics[width=\columnwidth]{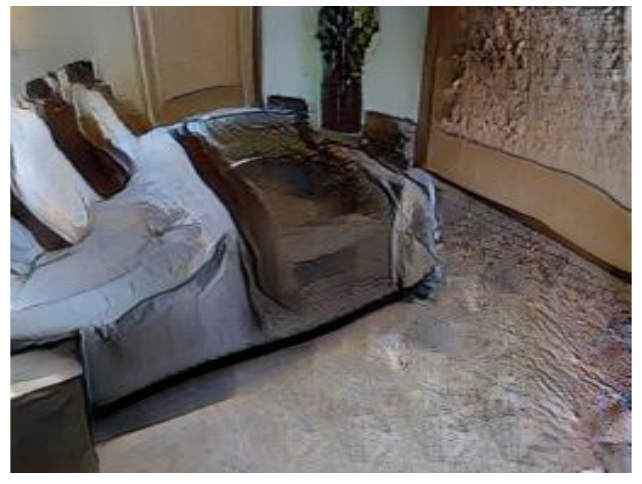} 
    \end{minipage} \\
   4& \begin{minipage}{.4\columnwidth}
      \includegraphics[width=\columnwidth]{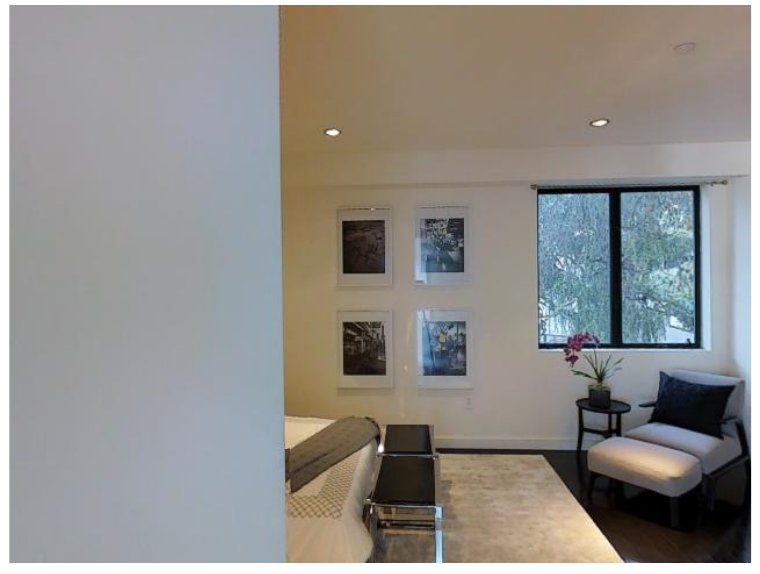} 
    \end{minipage} & \begin{minipage}{.4\columnwidth}
      \includegraphics[width=\columnwidth]{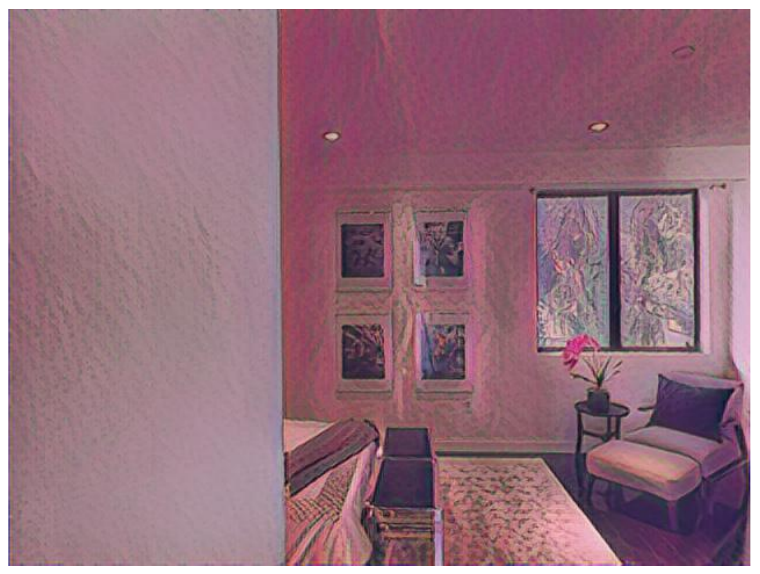} 
    \end{minipage} & \begin{minipage}{.4\columnwidth}
      \includegraphics[width=\columnwidth]{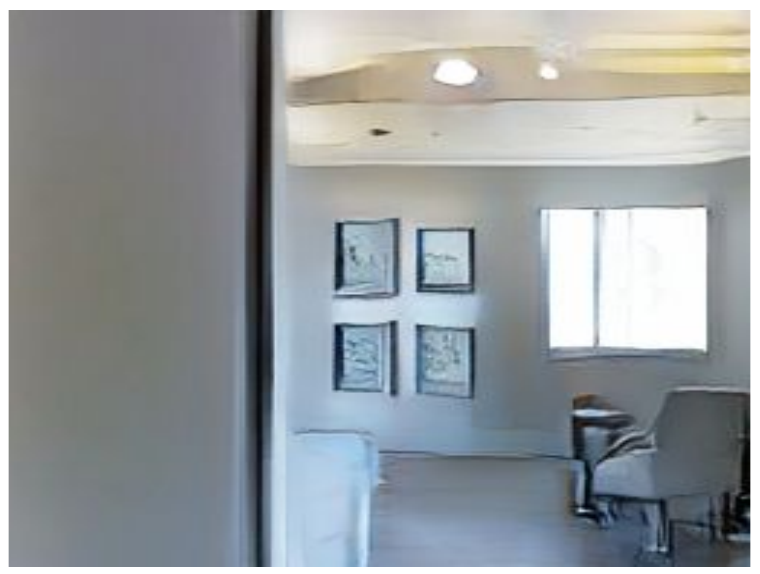} 
    \end{minipage}  & \begin{minipage}{.4\columnwidth}
      \includegraphics[width=\columnwidth]{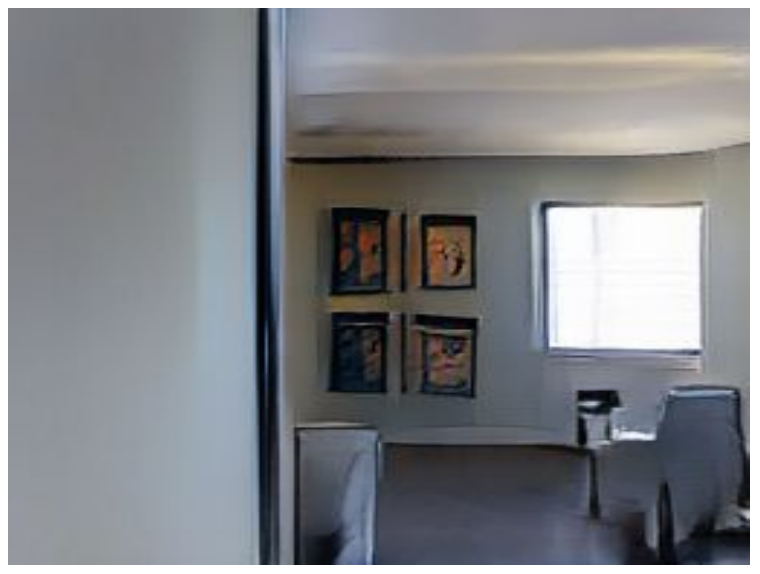} 
    \end{minipage} \\
   5 & \begin{minipage}{.4\columnwidth}
      \includegraphics[width=\columnwidth]{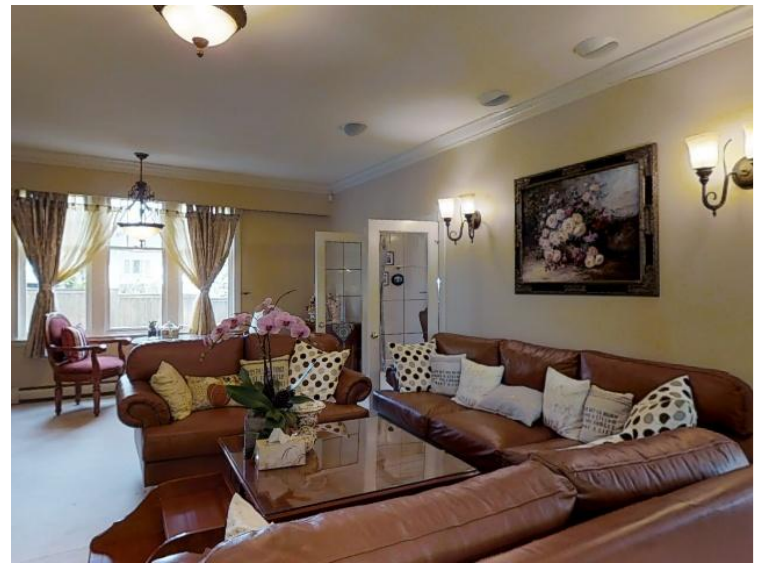} 
    \end{minipage} & \begin{minipage}{.4\columnwidth}
      \includegraphics[width=\columnwidth]{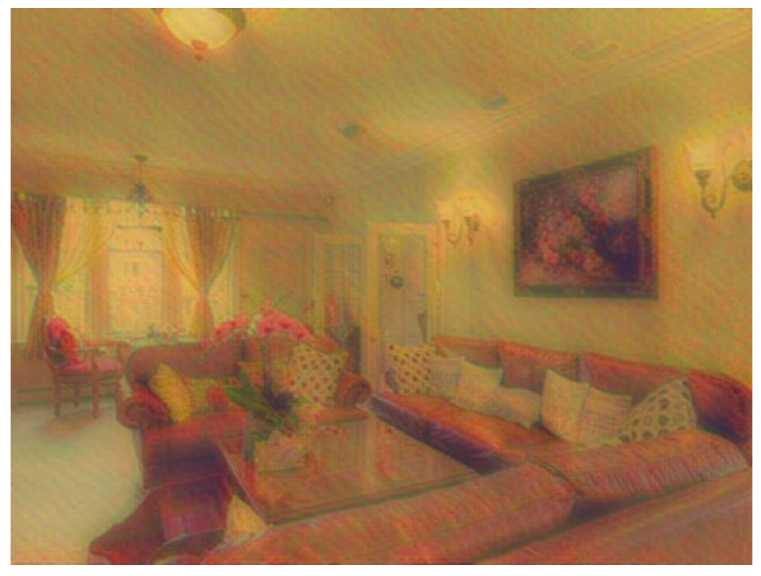} 
    \end{minipage} & \begin{minipage}{.4\columnwidth}
      \includegraphics[width=\columnwidth]{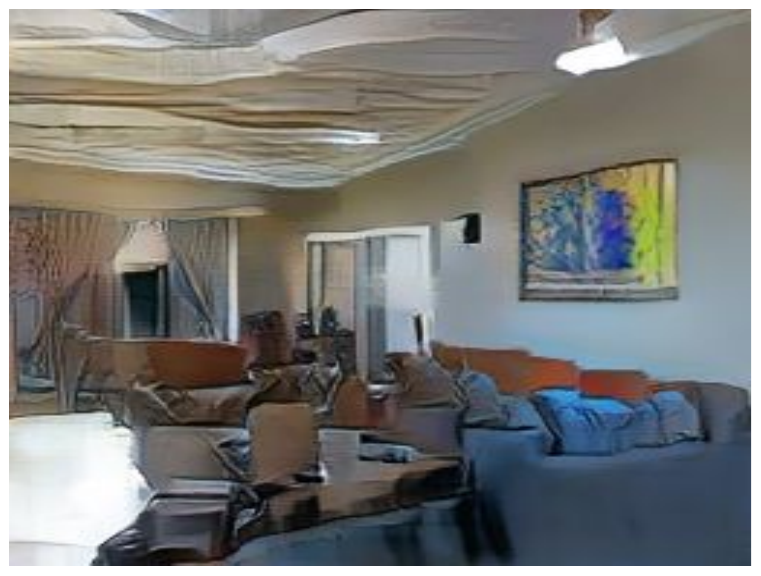} 
    \end{minipage}  & \begin{minipage}{.4\columnwidth}
      \includegraphics[width=\columnwidth]{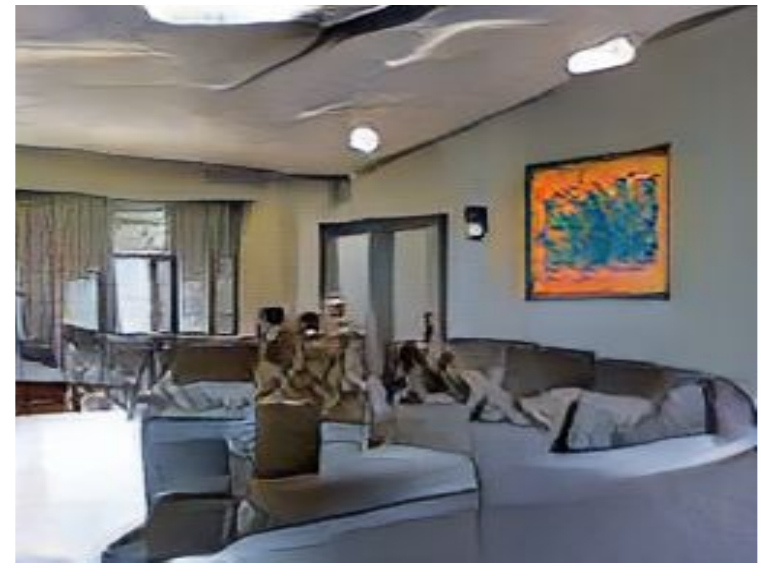} 
    \end{minipage} \\
     6 & \begin{minipage}{.4\columnwidth}
      \includegraphics[width=\columnwidth]{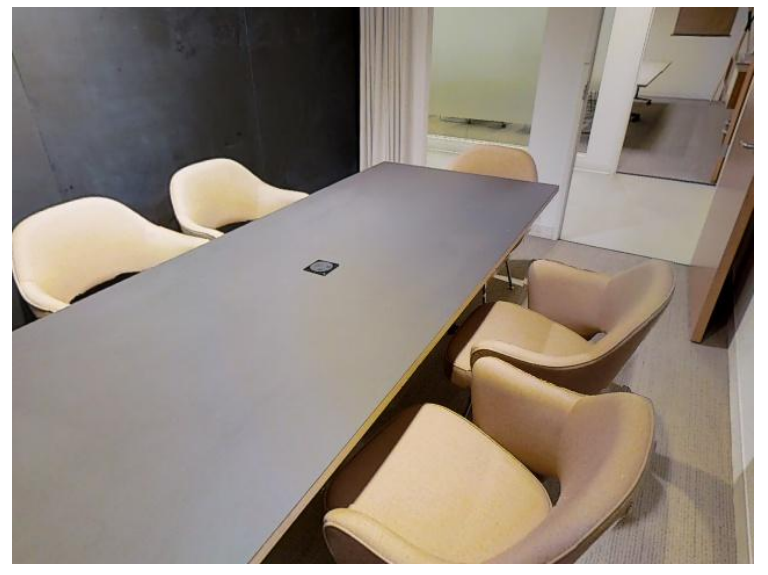} 
    \end{minipage} & \begin{minipage}{.4\columnwidth}
      \includegraphics[width=\columnwidth]{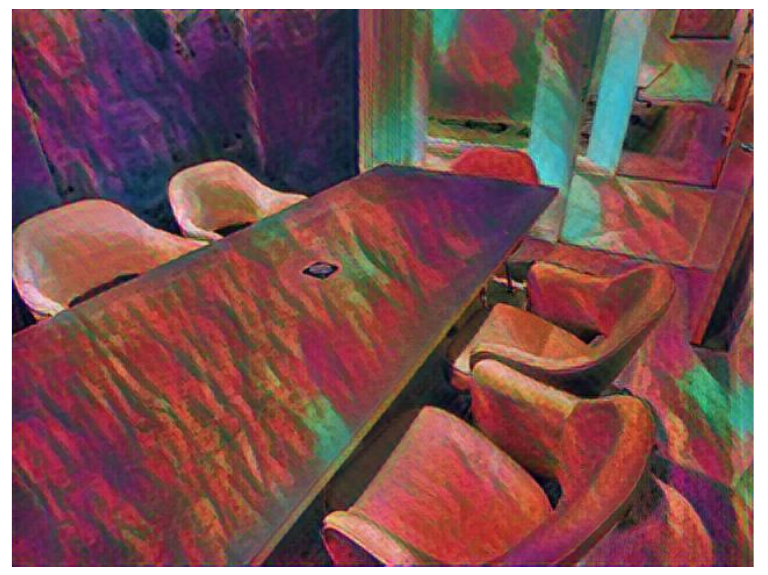} 
    \end{minipage} & \begin{minipage}{.4\columnwidth}
      \includegraphics[width=\columnwidth]{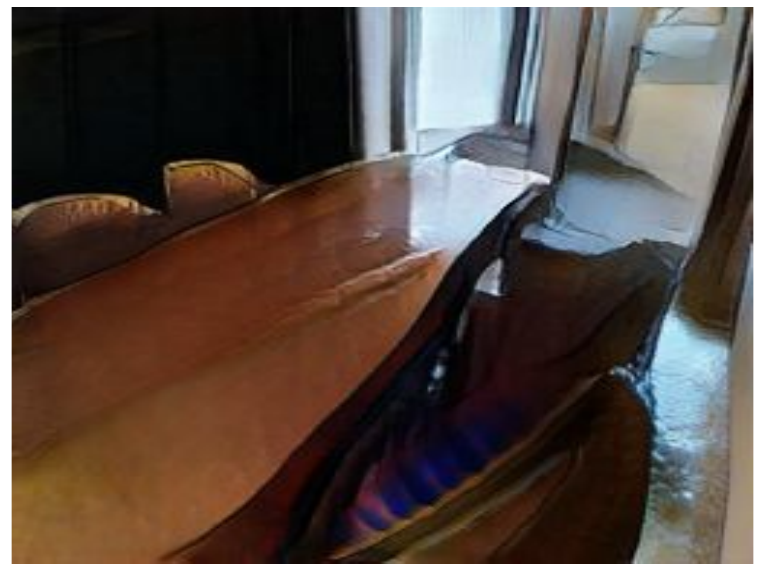} 
    \end{minipage}  & \begin{minipage}{.4\columnwidth}
      \includegraphics[width=\columnwidth]{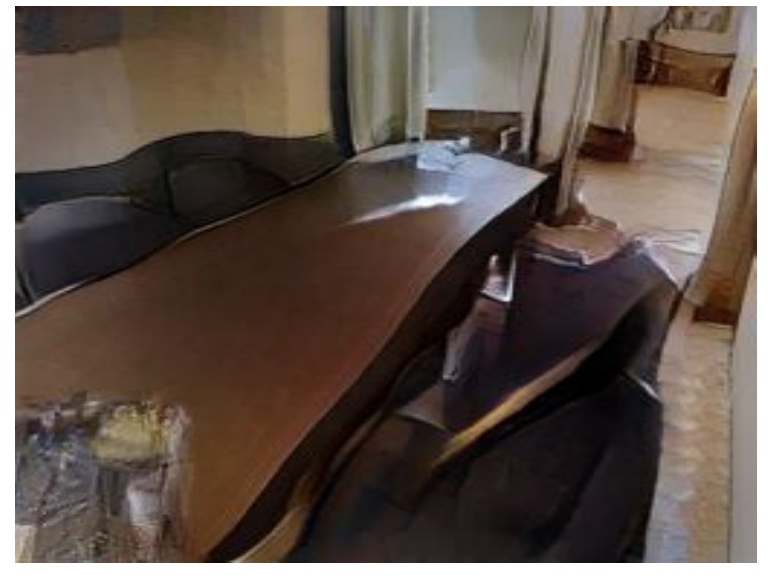} 
    \end{minipage} \\
    \hline
    \end{tabular}
    \caption{Qualitative Examples of our created environments $E_{st}$, $E_{is_1}$, $E_{is_1}^m$.}
    \label{table12_appendix}
\end{table*}

\section{Editing Different Number of Objects} \label{sec:spade}
We edit the objects in the original environments by randomly masking out some classes in the semantic segmentation (as discussed in Sec.4.2 in the main paper)
In this section, we explore the impact of editing different numbers of objects in the original environments.  Specifically, we randomly mask out 1 to 4 classes in the semantic segmentation during the inference time of environment editing, and create four environments respectively ($E_{is_1}^{m_1}$, $E_{is_1}^{m_2}$, $E_{is_1}^{m_3}$, $E_{is_1}^{m_4}$), where $E_{is_1}^{m_i}$ indicates masking out $i$ classes in the semantic segmentation. $E_{is_1}^{m_1}$ is the environment $E_{is_1}^m$ used in the main paper. 

As shown in Table~\ref{table10_appendix}, the performance consistently decreases while masking out more object classes in the environments. These results demonstrate the importance of balancing between the matches with the original instructions and the diversity of new environments, and we find that masking out 1 object class would empirically give the best results in our experimental setup.

\section{Detailed Results on Different Environments} \label{sec:eeval}
We show the results of the overall averaging performance on all unseen environments in the main paper. In this section, we want to have a detailed look at the performance of our model on different environments and explore which unseen environments benefit more from our \methodname{}. 
As shown in Table~\ref{table11_appendix}, we observe that training on $E_{st}$ could improve the performance on Environment $5$ by 15.0\% in SR and 13.8\% in SPL, and improve the performance on Environment $11$ by 7.2\% in SR and 4.6\% in SPL. 
When training on $E_{is_1}$, the Environment $1$ gets the largest improvement. A possible reason is that this environment differ from the original environment mainly in object appearance, and our \methodname{} creates environments with new object appearance, and thus generalize better.
Besides, this indicates that environments in the validation unseen set benefit differently from training on different created environments, and the newly created environments contain complementary information for generalization. 
This points to a future direction where the agent learns from multiple new environments to generalize better to unseen environments.

\section{Analysis of Consistency Across Different Environments} \label{sec:con}
In Sec~\ref{sec:eeval}, we show that different environments could benefit differently from different editing methods (i.e., Env5 gets 15\% improvement by $E_{st}$, while Env11 gets 38.9\% improvement by $E_{is_1}^m$). Similarly, in this section, we show that different VLN models (EnvDrop/EnvDrop+Back Translation(EnvDrop+BT)/$\circlearrowright$BERT-p365) will be better at generalization to different environments. As shown in Table~\ref{table13_appendix}, for VLN models trained on the original environment, while `EnvDrop' generalizes bad to Env5 (42.7\%), `EnvDrop+BT' improves it by 9\% (51.7\%) and `$\circlearrowright$BERT-p365' improves it by 12.6\% (55.3\%) on Env5. Thus, the improvement brought by $E_{st}$ might be less for `EnvDrop+BT' and `$\circlearrowright$BERT-p365' compared with `EnvDrop'. Similarly, $E_{is_1}^m$ can benefit `EnvDrop+BT' more since `EnvDrop+BT' still works bad on Env11 (22.2\%). This explains why the best model for each VLN model and visual feature will be different. Nevertheless, 
picking any of the editing methods will improve the performance, which demonstrates the effectiveness of our environment-level data augmentation in tackling the data scarcity problem. Considering both simplicity and performance across different models, we recommend using $E_{st}$ as the start point for future research.

\section{Qualitative Examples for Generated Environments} \label{sec:example}
We show some more examples in our generated environments in Table~\ref{table12_appendix}. We could see that the environments generated with the style transfer approach (denoted as $E_{st}$ in Table~\ref{table12_appendix}) maintain the semantics of the original environments better, but the overall style is artistic.
The environments generated with the image synthesis approach (denoted as $E_{is_1}$ and $E_{is_1}^m$ in Table~\ref{table12_appendix}) bring more diversity in object appearance and are more close to real environments.

\section{Limitations} \label{sec:limitation}
We also note that there are some limitations of our work. First, this work explores two classic style transfer and image synthesis approaches in augmenting the VLN training data. More advanced image-to-image translation models can potentially generate environments with higher quality and thus bring further improvement. Besides, this work dedicatedly considers the Vision-and-Language Navigation task, but the proposed method can be potentially used in other embodied tasks and simulated environments with certain adaptations. We will explore other useful and interesting tasks in the future.

\end{document}